\documentclass[11pt]{article}

\usepackage[final]{acl}

\usepackage{latexsym}
\usepackage{booktabs}
\usepackage{amssymb}
\usepackage{fontspec}
\usepackage[english]{babel}
\babelprovide[import]{arabic}
\babelprovide[import]{russian}
\babelprovide[import]{french}
\babelprovide[import]{spanish}
\usepackage{tabularx}
\usepackage{array}
\usepackage{makecell}
\usepackage{placeins}

\newfontfamily\cjkfont{FandolSong-Regular.otf}[BoldFont=FandolSong-Bold.otf]
\babelfont[arabic]{rm}[Script=Arabic,
  BoldFont       = Amiri-Bold.ttf,
  ItalicFont     = Amiri-Italic.ttf,
  BoldItalicFont = Amiri-BoldItalic.ttf]{Amiri-Regular.ttf}
\babelfont[russian]{rm}[Script=Cyrillic,
  BoldFont       = NotoSerif-Bold.ttf,
  ItalicFont     = NotoSerif-Italic.ttf,
  BoldItalicFont = NotoSerif-BoldItalic.ttf]{NotoSerif-Regular.ttf}

\usepackage{tikz}
\usepackage{pgfplots}
\usepackage{xcolor}
\pgfplotsset{compat=1.18}

\definecolor{bgeblue}{HTML}{2F5D8C}
\definecolor{e5green}{HTML}{3F7F5A}
\definecolor{labseorange}{HTML}{A65F1A}
\definecolor{accentred}{HTML}{9B1C1C}
\definecolor{contextgray}{HTML}{9A9A9A}

\usepackage{microtype}

\usepackage{inconsolata}

\usepackage{graphicx}

\usepackage{fvextra}
\DefineVerbatimEnvironment{PromptBlock}{Verbatim}{%
  fontsize=\footnotesize,
  breaklines=true,
  breakanywhere=true,
  breakindent=0pt,
  breakautoindent=false,
  breaksymbolleft={},
  breaksymbolright={},
  frame=single,
  framerule=0.35pt,
  framesep=1.5mm,
  rulecolor=\color{black!45},
  fillcolor=\color{black!2},
  xleftmargin=0pt,
  xrightmargin=0pt
}

\usepackage{bidi}
\newenvironment{arabicblock}%
  {\leavevmode\par\begin{otherlanguage*}{arabic}\begin{RTL}}%
  {\end{RTL}\end{otherlanguage*}}

\title{Do LLM Debates Repeat Arguments Differently Across Languages?}

\author{Huiqian Lai \\
  Syracuse University \\
  \texttt{hlai12@syr.edu}}

\begin{document}
\maketitle
\begin{abstract}
LLM debate is usually evaluated by final answers, yet transcripts reveal whether later turns develop new arguments or return to earlier claims in new wording. We study this process with \textit{prior-argument similarity}, which compares extracted argument units with earlier units in the same debate. In controlled eight-turn debates over 71 motions, six languages, and four model agents, Chinese is the only tested language with a consistently positive gap relative to English across three multilingual embedding models. The gap persists across agents, turn positions, regression adjustment, metric variants, extraction-length controls, a second-extractor subset, and cross-encoder tail rescoring. Manual calibration shows weak item-level alignment but a high-similarity tail enriched for substantive repetition. A diversity-aware prompt lowers \textit{prior-argument similarity} across languages, yet does not significantly narrow the Chinese--English gap. Multilingual debate evaluation should therefore measure argumentative development over time and report both average and gap terms.
\end{abstract}

\section{Introduction}

Multi-agent debate is often motivated by the promise that interaction can surface alternative perspectives, expose errors, and improve reasoning~\citep{du2023improving,chan2023chateval,yin2023exchange,zhang2024exploring}. Yet this promise depends on what happens inside the debate, not only on the final answer. Extended LLM interaction can also converge to shared misconceptions, stall, drift, or fail to produce genuinely useful exchange~\citep{liang2024encouraging,estornell2024multillmdebate,wynn2025talk,wu2025agentsreallydebate,becker2026stay}. This raises a process question beyond final-answer evaluation: \textbf{do later turns develop new argumentative content, or do they semantically return to prior content in new wording?}

Existing work does not directly answer this question. LLM debate is usually evaluated as an inference-time strategy for improving final answers, reasoning, or judgments~\citep{irving2018ai,du2023improving,liang2024encouraging,chan2023chateval}, while work that treats debate as an argumentative object focuses on judging, premise recovery, or counter-argument generation~\citep{liang2024debatrix,ku2025premise,ozaki2025debateopponent}. Multilingual and cross-cultural LLM studies show that model behavior is not language- or culture-neutral, but they largely evaluate final outputs or aggregate responses rather than turn-by-turn interaction dynamics~\citep{shi2023language,arora2023probing,naous2024having,gao2025multilingualreasoning}. Repetition research, in turn, studies degeneration in generated text at the lexical, phrase, or sentence level~\citep{holtzman2020curious,welleck2020neural,su2022learning}. Debate requires a different unit of analysis: a later turn can avoid copying earlier wording while still returning to the same claim, mechanism, stakeholder impact, or tradeoff.

We study argument development in multilingual, multi-agent LLM debate by extracting argument units from each turn and measuring \textit{prior-argument similarity}. For each extracted argument unit, we identify its most similar earlier unit in the same debate. Higher values mean that later speakers are returning to earlier argumentative ground, even when they use different wording. The score is a process diagnostic: it can reveal where debates exhibit greater semantic overlap with prior arguments, but it does not determine whether that overlap is quotation, rebuttal, refinement, or redundant restatement.

Our evaluation uses controlled eight-turn debates over 71 motions in the six official UN languages, which form a policy-relevant multilingual benchmark spanning distinct scripts and language families while remaining high-resource and supported by every debate agent and embedding model in the study; this reduces, without eliminating, the risk that observed differences merely reflect low-resource capability failure. Each session includes four provider-hosted model agents: OpenAI GPT-5.5~\citep{openai2026gpt55}, Anthropic Claude Opus 4.7~\citep{anthropic2026opus47}, DeepSeek-V3.2~\citep{deepseek2025v32}, and Mistral Large 3~\citep{mistral2025large3}. For each motion, the topic, role-to-agent assignment, debate positions, and turn structure are paired across languages, allowing language-conditioned comparisons while reducing avoidable variation from topic, role, model identity, and interaction structure.

This paper makes three contributions. \textit{First}, we introduce \textit{prior-argument similarity}, a scalable diagnostic for measuring whether later debate turns return to earlier argumentative content rather than developing new ground. \textit{Second}, using a yoked six-language design over 71 motions and four model agents, we show that Chinese debates exhibit consistently higher \textit{prior-argument similarity} than English debates across embedding models, agents, turn positions, regression adjustments, metric variants, and artifact checks. \textit{Third}, we triangulate this process diagnostic with human labels, a cross-encoder, a blinded LLM pair judge, a translate-and-rescore probe, prompt-strategy ablations, and an independent outcome-quality judge. The combined evidence supports \textit{prior-argument similarity} as an aggregate process monitor rather than an item-level repetition classifier or standalone quality score: higher similarity is moderately associated with lower judged quality, while prompt-based reductions in similarity do not necessarily yield equal downstream gains across languages.
\section{Related Work}
\label{sec:related}

\subsection{LLM Debate as Outcome Evaluation}
Most LLM debate work treats interaction as an inference-time route to a better final answer, score, or verdict. Agents propose, critique, and revise responses, and success is usually measured by downstream answer quality or judging accuracy~\citep{irving2018ai,du2023improving,liang2024encouraging,chan2023chateval}. Recent work broadens the setting to structured judging, code generation, cultural alignment, self-signal-driven debate, role-playing debate benchmarks, and applied debate protocols~\citep{harrasse2026d3,chen2025debatecoder,ki2025cultural,chen2025sid,chuang2025debate,oriol2025requirementsdebate}. Stress tests of debate architectures further show that debate can converge to shared misconceptions, underperform single-agent reasoning, or become vulnerable to social and persuasive influence~\citep{estornell2024multillmdebate,wynn2025talk,wu2025agentsreallydebate,kraidia2026collaborationfails}. Across these lines, the central evaluation object is still usually the outcome of the interaction.

A smaller body of work treats debate as an argumentative object. Debate judging requires tracking how claims are introduced, supported, and refuted across turns~\citep{liang2024debatrix,sternlicht2025debatable}; multi-agent debate can surface implicit premises~\citep{ku2025premise}; counter-argument generation depends on identifying critical premises~\citep{ozaki2025debateopponent}; and parliamentary debate summarization emphasizes preservation of reasoning structure~\citep{cunningham2026parliamentary}. Debate-style latent-role methods such as MoLaCE similarly aim to increase reasoning diversity~\citep{kim2025molace}. These studies motivate argument-level analysis, but do not measure whether later turns expand the argumentative state or return to earlier ground.

\subsection{Multilingual LLM Evaluation Beyond Final Outputs}
Multilingual evaluation has consistently shown that LLM behavior is not language-neutral~\citep{shi2023language,arora2023probing,naous2024having,gao2025multilingualreasoning}. Parallel and translated benchmarks reveal cross-lingual variation in reasoning, instruction following, safety, and cultural alignment~\citep{huang2024m3exam,gao2025multilingualreasoning}. Prompting can reduce some of these gaps, as in cross-lingual-thought prompting for multilingual reasoning and generation~\citep{huang2023notall}. This work establishes language-conditioned variation, but it mostly measures final task scores or aggregate responses.

Recent work also cautions that multilingual and cultural evaluations are sensitive to format. Multiple-choice evaluations can be order-sensitive and can diverge from long-form generation in output, logit, and embedding spaces~\citep{li2024mcq}; closed-style cultural-alignment surveys can obscure behavior that appears under less constrained generation~\citep{kabir2025break}. These findings are directly relevant to multilingual debate: if format changes measured behavior in single-turn settings, then interactive systems should not be reduced to final-answer accuracy alone. What remains missing is a yoked, turn-level account of multilingual interaction: whether the same motions, roles, agents, and turn positions produce comparable argumentative development across languages.

\subsection{From Surface Repetition to Argument-Level Repetition}
Repetition is a well-studied problem in neural text generation. Prior work analyzes degenerate continuations and mitigates repeated output with sampling, diverse decoding, unlikelihood training, repetition penalties, and preference-based fine-tuning~\citep{holtzman2020curious,welleck2020neural,vijayakumar2017diverse,su2022learning,wang2025repetition}. Evaluation work also shows that repeated generations can be unstable across runs, motivating repeated sampling or aggregation~\citep{vardasbi2025adaptive,alvarado2025repetitions}. These lines target surface or generation-level repetition; debate requires a different object, because a turn may avoid lexical repetition while returning to the same claim, mechanism, stakeholder impact, or tradeoff.

Argument mining and semantic similarity work provide the needed unit of analysis. The same substantive argument can appear in different wording, motivating argument similarity, prominent-argument identification, argument facets, and contextualized argument clustering~\citep{boltuzic2015identifying,misra2016measuring,reimers2019classification}. Sentence-pair modeling and Sentence-BERT further distinguish accurate cross-encoder comparison from efficient bi-encoder retrieval and clustering~\citep{reimers2019sentencebert}. Diversity and redundancy metrics in dialogue, retrieval, and summarization are also relevant, from diversity-promoting dialogue objectives to maximal marginal relevance~\citep{li2016diversity,carbonell1998mmr}. Our setting differs because the units are extracted arguments, the comparison is within a structured debate transcript, and the target is a process-level diagnostic of semantic return, calibrated against human judgments and tested under matched multilingual intervention conditions.
\section{Methodology}
\label{sec:method}

\subsection{Experimental Design}
\label{sec:experimental-design}
We study semantic return to prior argumentative content in multilingual LLM debate, operationalized as prior-argument similarity to arguments already present in the same debate. Each debate session is defined by a motion, a language, and a prompt condition. The baseline experiment contains 426 sessions: 71 motions $\times$ six languages (English, Arabic, Spanish, French, Russian, and Chinese). These six official UN languages span multiple scripts and language families while remaining high-resource and supported by every tested API and embedding system, which reduces a low-resource capability confound. A matched diversity-aware condition repeats the same grid with an additional novelty instruction.

The motions are drawn from the World Universities Debating Championship (WUDC) motion archive for the 2023--2026 tournaments~\citep{tokyodebate_wudc_motions}. We translated each English motion into five target languages (355 cells) with Google Translate, back-translate with \texttt{deepseek-chat}, and judge meaning preservation with Llama-4-Scout before manual triage. Flagged cases were retained in the primary analysis and evaluated through exclusion-based sensitivity checks; Appendix~\ref{app:translation-qa} reports all parameters, schemas, flag rules, and corrected counts.

Each session is an eight-turn structured debate with two sides, four debate roles, and four provider-hosted model agents: OpenAI GPT-5.5~\citep{openai2026gpt55}, Anthropic Claude Opus 4.7~\citep{anthropic2026opus47}, DeepSeek-V3.2~\citep{deepseek2025v32}, and Mistral Large 3~\citep{mistral2025large3}. For each motion, the role-to-model assignment is fixed across all six language versions. This yoked design separates language from topic, role, turn structure, or model identity. Appendix Figure~\ref{fig:debate-process} schematizes the debate-generation protocol; detailed model identifiers, the debate template, execution counts, and operational notes are in Appendix~\ref{app:method-details}.

At each turn, the active agent receives the motion, its side and role, the current turn label, the immediately preceding speech, and the debate history so far. The agent is instructed to write only in the target language. To reduce length-driven variation, opening, rebuttal, and cross-examination turns contain exactly five sentences, while closing turns contain exactly four sentences; each sentence is constrained to be under 50 words. Appendix~\ref{app:baseline-prompt} gives the full generation prompt.

\paragraph{Diversity-aware prompt.}
The diversity-aware condition uses the same motions, languages, agents, role assignments, and turn order as the baseline condition. It appends one translated instruction asking the active agent to introduce at least one new argument, example, causal mechanism, stakeholder impact, or tradeoff, and to extend rather than merely restate any earlier point. The exact wording is in Appendix~\ref{app:diversity-prompt}.

\subsection{Argument Extraction}
\label{sec:argument-extraction}
We analyze debates at the argument-unit level rather than the full-turn level. After generation, each turn is decomposed into one to four argument units. An argument unit is a distinct substantive claim or reason made by the speaker; such units may express causal mechanisms, examples, stakeholder impacts, tradeoffs, or evaluative points.

Argument extraction is performed with Llama-4-Scout-17B-16E-Instruct at temperature~0. The extractor returns strict JSON fields for \texttt{claim}, \texttt{warrant}, \texttt{evidence\_or\_example}, and \texttt{argument\_type}. The \texttt{argument\_type} label is used only for argument-category diagnostics, not for prior-argument similarity. The extraction prompt, schema, deployment details, parse-failure rate, and diagnostics are in Appendix~\ref{app:argument-schema}.

\subsection{Prior-Argument Similarity}
\label{sec:prior-argument-similarity}
Our main dependent variable is \texttt{mean\_arg\_max\_prev}, a turn-level prior-argument similarity score. Higher values indicate that the current turn's extracted claims are more semantically similar to claims from earlier turns in the same debate. We use this score as an aggregate diagnostic of semantic return to prior argumentative content, not as a direct human judgment of repetitiveness or debate quality. Because the score takes a maximum over prior arguments, it identifies the closest earlier overlap; it does not itself distinguish quotation, rebuttal, refinement, or productive reuse from redundant restatement.

For the primary specification, we embed the \texttt{claim} field of each extracted argument unit. For an argument produced at turn~$t$, we compare its embedding to the embeddings of all argument units produced in strictly earlier turns of the same session, take the maximum cosine similarity, and then average these maxima across the current turn's argument units.

Formally, fix a debate session~$s$. Let $a_{s,t,i}$ denote the $i$-th extracted argument unit in turn~$t$, and let $c_{s,t,i}$ denote its \texttt{claim}. For embedding model~$e$, let $z_e(c_{s,t,i}) \in \mathbb{R}^{d_e}$ be the L2-normalized embedding of that claim. The prior pool for turn~$t$ is
\[
P_{s,t}=\{a_{s,t',j}:1 \leq t' < t,\; 1 \leq j \leq n_{s,t'}\},
\]
where $n_{s,t'}$ is the number of extracted arguments in turn~$t'$. The per-argument prior-similarity score is
\[
r_e(a_{s,t,i})=
\max_{a_{s,t',j}\in P_{s,t}}
z_e(c_{s,t,i})^\top z_e(c_{s,t',j}).
\]
Because embeddings are L2-normalized, the dot product is cosine similarity. The turn-level score is
\[
\texttt{mean\_arg\_max\_prev}^{(e)}_{s,t}
=
\frac{1}{n_{s,t}}
\sum_{i=1}^{n_{s,t}} r_e(a_{s,t,i}).
\]
The score is defined only for turns with a non-empty prior pool, excluding turn~1.

We compute the metric with three multilingual sentence embedding models: BGE-M3~\citep{chen2024bgem3}, multilingual-E5-large~\citep{wang2024multilinguale5}, and LaBSE~\citep{feng2022labse}. BGE-M3 is the primary specification for narrative discussion; multilingual-E5-large and LaBSE are used as independent representation checks. Similarity is always computed within a single language session. The analysis therefore does not directly compare Chinese claims to English claims; it compares within-session prior-similarity levels after aggregating by language.

\subsection{Metric Calibration, Diagnostic Variants, and Manual Audit}
\label{sec:metric-diagnostics}

We include diagnostics for possible extraction and similarity artifacts. These include language-specific null calibration, $z$-scoring, fixed-size prior pools, input-field ablations, extraction diagnostics, and sentence-count compliance checks. Together, they test whether measured language gaps are plausibly driven by embedding calibration, pool size, input-field choice, extraction granularity, or sentence-template compliance.

We additionally conduct a bilingual English--Chinese manual audit. Annotators evaluate extraction quality and label sampled current--prior argument pairs as new, topically related, partial repetition or extension, or substantive repetition. The same 200 adjudicated pairs are then scored by a multilingual cross-encoder and by a blinded \texttt{deepseek-chat} pair judge, which serve as complementary ranking instruments. Together these analyses calibrate the diagnostic rather than converting it into a direct perceptual repetition label; protocols, agreement, and results are in Appendix~\ref{app:manual-audit-protocol}.

\subsection{Outcome-Quality Evaluation}
\label{sec:outcome-quality-method}
Calibration tells us what the score measures, but not whether it tracks debate quality. We therefore add an independent outcome measure: a blinded Gemini-3.5-Flash judge, which is not one of the four debate agents, scores all 284 English and Chinese baseline and diversity sessions on argumentative development, engagement, final argument quality, and overall quality, with 20 sessions rescored for stability. Appendix~\ref{app:outcome-quality} gives the protocol and analyses.

\subsection{Statistical Inference}
\label{sec:stats}
All language contrasts are paired by motion. For each non-English language~$\ell$, motion~$m$, embedding model~$e$, and metric~$M$, let $M^{(e)}(\ell,m)$ denote the session-level metric for language~$\ell$ and motion~$m$, averaged across model agents when analyses are pooled. We compute
\[
d^{(e)}_{\ell,m}=M^{(e)}(\ell,m)-M^{(e)}(\mathrm{en},m).
\]
Pairing on motion controls for topic-level variation in how naturally a motion invites repetition. Agent-level analyses use the analogous paired contrast within each model agent.

Uncertainty is estimated with a cluster bootstrap that resamples motions with replacement. Unless otherwise stated, reported confidence intervals use percentile intervals from fixed-seed bootstrap replicates, and a contrast is treated as statistically distinguishable from zero when the 95\% interval excludes zero. As an additional robustness check for the focal Chinese--English contrast, we fit turn-level OLS models with motion fixed effects and motion-clustered standard errors, adjusting for turn label, model agent, role, and side. Additional bootstrap and regression diagnostics, including motion-random-intercept mixed models, are in Appendices~\ref{app:bootstrap} and~\ref{app:regression-robustness}.

\subsection{Robustness and Intervention Analyses}
\label{sec:robustness}
The main analysis first estimates the six-language landscape relative to English, then applies robustness and intervention analyses to the non-English language whose gap is sign-consistent across all three embedding models. Detailed procedures for agent, turn, metric, artifact, decoding, translation-QA, intervention, and repeat-generation analyses are in Appendix~\ref{app:robustness-diagnostics}. Two further checks reuse the same locked pipeline: a six-condition, three-repeat prompt ablation on 10 English--Chinese motions, which asks whether the contrast is an artifact of one prompt wording, and a 30-motion Chinese$\rightarrow$English translate-and-rescore with an English round-trip control, which separates a surface-language component from a content-level one. Both test consistency and partial mechanism decomposition, not complete causal identification.
\section{Results}
\label{sec:results}

We first scan six languages, then test the stable contrast against design, metric, and artifact explanations; calibrate the score with human and model-based instruments; evaluate prompt interventions and outcome quality; and check repeat-generation stability.

Unless otherwise stated, language contrasts are paired by motion and use a 5{,}000-sample motion-cluster bootstrap with percentile confidence intervals. The outcome is \texttt{mean\_arg\_max\_prev}: each turn's average maximum cosine similarity to earlier argument units in the same debate. Higher values indicate stronger semantic return to prior argumentative content, so we interpret the score as an aggregate diagnostic rather than an item-level repetition label.

\subsection{A six-language scan isolates Chinese as the sign-consistent gap}
\label{sec:res-crosslingual}

\begin{figure*}[t]
\centering
\begin{tikzpicture}
\begin{axis}[
    width=0.94\textwidth,
    height=0.46\textwidth,
    xmin=-0.04,
    xmax=0.10,
    ymin=0.55,
    ymax=5.45,
    xlabel={$\Delta$ prior-argument similarity (language -- English)},
    xlabel style={font=\small},
    ytick={1,2,3,4,5},
    yticklabels={Arabic,Spanish,French,Russian,\textbf{\textcolor{accentred}{Chinese}}},
    y dir=reverse,
    axis x line*=bottom,
    axis y line*=left,
    tick align=outside,
    tick style={black!60},
    xtick={-0.04,-0.02,0.00,0.02,0.04,0.06,0.08,0.10},
    scaled x ticks=false,
    xticklabel style={
        /pgf/number format/fixed,
        /pgf/number format/precision=2,
        font=\small
    },
    yticklabel style={font=\small},
    xmajorgrids=true,
    ymajorgrids=false,
    major grid style={dotted, gray!35},
    legend style={
    draw=none,
    fill=white,
    fill opacity=0.88,
    text opacity=1,
    font=\small,
    at={(1.05,0.98)},
    anchor=north east,
    row sep=1pt
},
    legend cell align=left,
    clip=false
]

\addplot[
    black!55,
    line width=0.7pt,
    forget plot
] coordinates {(0,0.55) (0,5.45)};

\addplot[
    only marks,
    mark=*,
    mark size=2.8pt,
    color=bgeblue,
    error bars/.cd,
    x dir=both,
    x explicit,
    error bar style={line width=0.75pt, bgeblue},
    error mark options={rotate=90, mark size=1.7pt, line width=0.75pt}
]
table[
    x=estimate,
    y=y,
    x error minus=errminus,
    x error plus=errplus
]{
estimate y errminus errplus
-0.024599 0.82 0.007127 0.006734
-0.021702 1.82 0.006233 0.006200
-0.017468 2.82 0.005456 0.005285
-0.009617 3.82 0.006209 0.006054
 0.046861 4.82 0.006431 0.006211
};
\addlegendentry{BGE-M3}

\addplot[
    only marks,
    mark=square*,
    mark size=2.6pt,
    color=e5green,
    error bars/.cd,
    x dir=both,
    x explicit,
    error bar style={line width=0.75pt, e5green},
    error mark options={rotate=90, mark size=1.7pt, line width=0.75pt}
]
table[
    x=estimate,
    y=y,
    x error minus=errminus,
    x error plus=errplus
]{
estimate y errminus errplus
0.038565 1.00 0.002463 0.002427
0.025714 2.00 0.002305 0.002304
0.027247 3.00 0.002129 0.002092
0.020643 4.00 0.002364 0.002168
0.052546 5.00 0.002030 0.002079
};
\addlegendentry{multilingual-E5-large}

\addplot[
    only marks,
    mark=triangle*,
    mark size=3.0pt,
    color=labseorange,
    error bars/.cd,
    x dir=both,
    x explicit,
    error bar style={line width=0.75pt, labseorange},
    error mark options={rotate=90, mark size=1.7pt, line width=0.75pt}
]
table[
    x=estimate,
    y=y,
    x error minus=errminus,
    x error plus=errplus
]{
estimate y errminus errplus
0.028687 1.18 0.010387 0.010498
0.057714 2.18 0.007411 0.007563
0.065212 3.18 0.008143 0.008255
0.083267 4.18 0.008068 0.008163
0.046580 5.18 0.009188 0.008920
};
\addlegendentry{LaBSE}

\end{axis}
\end{tikzpicture}
\caption{Cross-lingual prior-argument-similarity gaps relative to English.}
\label{fig:crosslingual-landscape}
\parbox{\linewidth}{\raggedright\footnotesize \textit{Note.} Points show differences in \texttt{mean\_arg\_max\_prev}; horizontal bars show 95\% motion-cluster bootstrap confidence intervals. Chinese is the only non-English language with a positive gap across all three embedding models.}
\end{figure*}
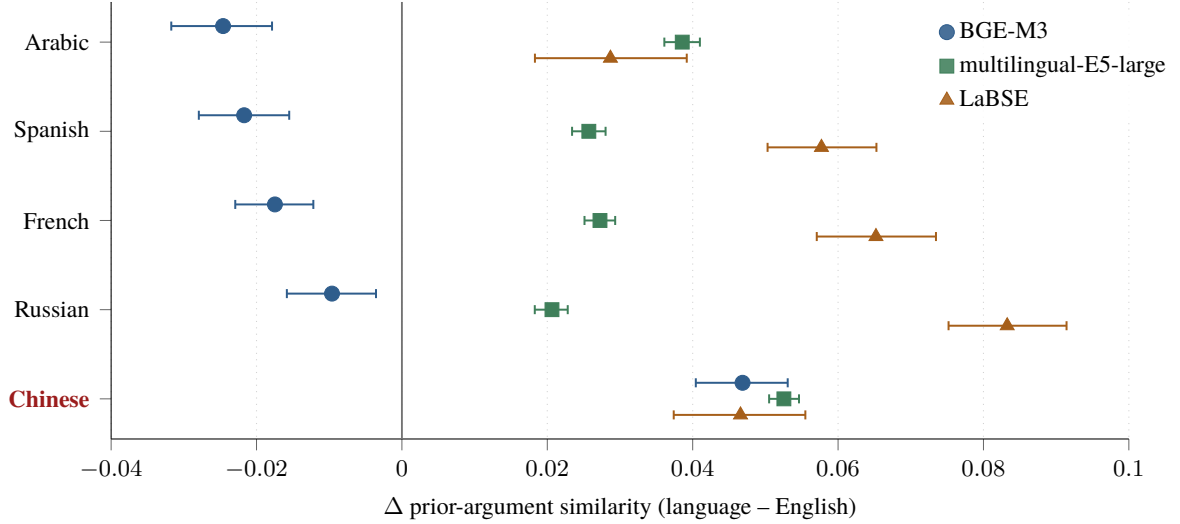

The full six-language comparison does not show a uniform non-English effect. In Figure~\ref{fig:crosslingual-landscape}, Arabic, Spanish, French, and Russian change direction across embedding models: they are below English under BGE-M3, but above English under multilingual-E5-large and LaBSE. Chinese is the exception. Its gap relative to English is positive under all three embedding models: $+0.047$ with BGE-M3, $+0.053$ with multilingual-E5-large, and $+0.047$ with LaBSE, with all 95\% motion-cluster bootstrap confidence intervals excluding zero. We therefore use Chinese--English as the focal contrast; exact six-language estimates are reported in Appendix Table~\ref{tab:app-lang-gaps}.

\subsection{Robustness checks narrow the plausible explanations}
\label{sec:res-robustness}

The following checks do not identify a mechanism, but they make the most direct confound explanations less plausible.

\paragraph{Design checks.}
The contrast is not concentrated in a single model agent or debate phase. Across four model agents and three embeddings, all $12$ model--embedding cells have a positive Chinese--English gap. Under BGE-M3, the model-agent gaps range from $+0.034$ for GPT-5.5 to $+0.061$ for Mistral Large 3, so agent identity modulates magnitude but not sign; Mistral's gap exceeds GPT-5.5's under all three embeddings. With one closed model per provider, size and family effects are not identifiable (Appendix Tables~\ref{tab:agent-language-gaps-full} and~\ref{tab:agent-pairwise-focal}). The turn-level trajectory also stays positive throughout the debate: prior-argument similarity rises as the prior pool grows, but Chinese remains above English from early constructive turns through closing. Regression adjustment gives the same conclusion. On baseline English--Chinese turns with non-empty prior pools ($N=994$), OLS models with motion fixed effects, motion-clustered standard errors, and controls for turn label, model agent, role, and side estimate positive Chinese coefficients under all three embeddings ($+0.0464$, $+0.0529$, and $+0.0469$ for BGE-M3, multilingual-E5-large, and LaBSE), consistent with the bootstrap estimates. Supporting agent, turn, and regression diagnostics are in Appendix Figures~\ref{fig:zh-en-robustness}--\ref{fig:turn-trajectory} and Appendix Table~\ref{tab:zhen-regression-robustness}.

\paragraph{Metric checks.}
Metric variants preserve the main contrast while clarifying its limits. In Table~\ref{tab:metric-diagnostics}, the gap remains positive under fixed-size prior pools ($K=5$ and $K=10$) and when embedding \texttt{claim+warrant} or \texttt{claim+warrant+evidence\_or\_example}, so it is unlikely to be only an artifact of a growing prior pool or a claim-only representation. Null calibration is more sensitive: the null-adjusted gap stays positive under BGE-M3 and LaBSE, but reverses under multilingual-E5-large. The corresponding $z$-scored variants are positive under all three embeddings, and $z$ calibration leaves Chinese as the only positive language. This is consistent with the embeddings differing in absolute offsets rather than in what they rank high or low: within-language turn-level rank agreement is high ($\rho=0.70$--$0.83$; Appendix~\ref{app:embedding-calibration}). We therefore read calibrated variants as representation-geometry sensitivity checks, not as evidence that every normalization gives the same absolute estimate. As an orthogonal tail check, rescoring all English--Chinese BGE-M3 max-prior pairs with a multilingual cross-encoder reranker preserves Chinese overrepresentation in the top-decile tail (ZH--EN $+0.048$), although with smaller effects than BGE-M3 (Appendix Table~\ref{tab:cross-encoder-tail}).

\begin{table}[t]
\centering
\small
\setlength{\tabcolsep}{3pt}
\begin{tabular}{lccc}
\toprule
Variant & BGE-M3 & mE5-large & LaBSE \\
\midrule
Original & $+0.0469^{*}$ & $+0.0525^{*}$ & $+0.0466^{*}$ \\
Null-adjusted & $+0.0147^{*}$ & $-0.0075^{*}$ & $+0.0343^{*}$ \\
$z$-scored & $+0.5171^{*}$ & $+1.4964^{*}$ & $+0.3933^{*}$ \\
Fixed $K=5$ & $+0.0494^{*}$ & $+0.0588^{*}$ & $+0.0452^{*}$ \\
Fixed $K=10$ & $+0.0432^{*}$ & $+0.0522^{*}$ & $+0.0418^{*}$ \\
Claim+W & $+0.0594^{*}$ & $+0.0487^{*}$ & $+0.0437^{*}$ \\
Claim+W+evidence & $+0.0592^{*}$ & $+0.0474^{*}$ & $+0.0438^{*}$ \\
\bottomrule
\end{tabular}
\caption{Metric diagnostics for the Chinese--English gap.}
\label{tab:metric-diagnostics}
\par\vspace{2pt}
\parbox{\linewidth}{\raggedright\footnotesize \textit{Note.} Positive values indicate higher prior-argument similarity in Chinese than English. Asterisks mark 95\% motion-cluster bootstrap intervals excluding zero. The multilingual-E5-large null-adjusted variant is the only sign reversal.}
\end{table}

\paragraph{Artifact checks.}
The contrast does not disappear under immediate artifact checks. Baseline argument counts are nearly flat across languages ($3.82$--$3.92$ arguments per turn), and Chinese does not concentrate disproportionately in the largest argument categories. Sentence-template compliance does not single out Chinese, decoding diagnostics do not remove the contrast, and conservative translation-QA exclusions move the point estimates by at most $0.0023$. The gap is also not removed by controls for claim length, empty warrant and evidence rates, argument count, prior-pool size, sentence count, turn label, model agent, role, side, and motion fixed effects (BGE-M3 adjusted coefficient $+0.0431$). Finally, a targeted 30-motion English--Chinese subset check with DeepSeek-Chat as a second extractor preserves the BGE-M3 gap ($+0.0506$ vs. $+0.0472$ under the original extractor), even though the second extractor nearly eliminates the empty-warrant imbalance. Detailed extraction, sentence-compliance, decoding, translation, extraction-control, and second-extractor diagnostics are reported in Appendix Tables~\ref{tab:extraction-diagnostics}, \ref{tab:sentence-compliance-summary}, \ref{tab:translation-sensitivity}, \ref{tab:zhen-extraction-controls}, and~\ref{tab:second-extractor-robustness}, and in Appendix~\ref{app:decoding-diagnostics}.

\paragraph{Surface language versus content.}
We then ask how much of the remaining gap travels with the surface language. Translating Chinese sessions into English and rescoring them in English embedding space removes most of the native BGE-M3 gap ($+0.047$ to $+0.009$), implicating surface form or representation geometry, but a smaller component survives an English round-trip control under all three embeddings ($+0.014$ under BGE-M3). This is a partial decomposition, not a causal account: it does not identify where that residual originates (Appendix~\ref{app:translate-rescore}).

\subsection{Human calibration localizes the signal to a repetition-enriched tail}
\label{sec:res-manual-calibration}

The manual audit narrows what the automatic score can support. Table~\ref{tab:manual-calibration} summarizes the main calibration statistics; Appendix Tables~\ref{tab:manual-pop-weighted} and~\ref{tab:manual-global-tail} and Appendix~\ref{app:manual-repetition-calibration} give the weighted labeled rates, the full tail breakdown, and a qualitative example.

\begin{table}[t]
\centering
\small
\setlength{\tabcolsep}{3pt}
\begin{tabularx}{\columnwidth}{@{}>{\raggedright\arraybackslash}Xcc@{}}
\toprule
Measure & Estimate & 95\% CI \\
\midrule
Extraction quality (ZH--EN) & $-0.06$ & $[-0.17,+0.04]$ \\
Exact agreement & $0.83$ & -- \\
Binary $\kappa$ & $0.733$ & -- \\
Weighted $\kappa$ & $0.857$ & -- \\
Spearman $\rho$ & $0.195$ & -- \\
AUC & $0.591$ & -- \\
Top-quintile rate (ZH--EN) & $+0.176$ & $[+0.148,+0.204]$ \\
Top-decile rate (ZH--EN) & $+0.074$ & $[+0.051,+0.096]$ \\
Calibrated strict-rep. gap & $+0.053$ & $[+0.015,+0.089]$ \\
\bottomrule
\end{tabularx}
\caption{Manual calibration on the English--Chinese audit subset.}
\label{tab:manual-calibration}
\par\vspace{2pt}
\parbox{\linewidth}{\raggedright\footnotesize \textit{Note.} ZH--EN denotes Chinese minus English.}
\end{table}

\paragraph{Item-level calibration.}
Extraction quality is comparable in the English--Chinese audit (ZH--EN $-0.06$, 95\% CI $[-0.17,+0.04]$). At the level of individual pairs, however, the automatic score aligns only weakly with human repetition labels: BGE-M3 reaches $\rho=0.195$, a cross-encoder improves modestly to $0.255$, and a blinded LLM pair judge improves ranking alignment to $0.553$ but not label calibration. No instrument closes the gap, which points to a construct difference as well as scorer limits: a pair can be near-duplicate in content while functioning as a rebuttal (Appendix Tables~\ref{tab:instrument-ladder} and~\ref{tab:high-sim-nonredundant}).

\paragraph{Tail calibration.}
The aggregate distribution is clearer. Under global BGE-M3 thresholds, Chinese argument units are more likely than English argument units to fall in the top quintile ($0.288$ vs. $0.112$; ZH--EN $+0.176$, 95\% CI $[+0.148,+0.204]$) and top decile ($0.137$ vs. $0.063$; ZH--EN $+0.074$, 95\% CI $[+0.051,+0.096]$). Combining these population tail rates with the pooled human calibration curve yields a calibrated strict-repetition gap of $+0.053$ (95\% CI $[+0.015,+0.089]$; Appendix Table~\ref{tab:manual-global-tail}). Because the direct labeled subset is too small and stratified to estimate the language gap on its own, this is tail-sensitive evidence about high-overlap argumentative content. An exploratory functional decomposition of sampled top-decile pairs confirms that the tail is mixed, containing redundant restatement as well as rebuttal and refinement, so we keep the interpretation at the level of an aggregate semantic-return diagnostic (Appendix Table~\ref{tab:function-label-distribution}).

\subsection{Asking for new arguments lowers similarity but does not close the gap}
\label{sec:res-intervention}

\begin{figure}[t]
\centering

\pgfplotslegendfromname{fig:diversity-legend-single}

\vspace{0.15em}

\begin{tikzpicture}
\begin{axis}[
    width=\linewidth,
    height=4.45cm,
    xmin=-0.040,
    xmax=0.000,
    ymin=0.45,
    ymax=6.55,
    title={(A) Prompt effect within each language},
    title style={font=\small},
    xlabel={Prompt $-$ baseline},
    xlabel style={font=\small},
    ytick={1,2,3,4,5,6},
    yticklabels={English,Arabic,Spanish,French,Russian,Chinese},
    y dir=reverse,
    axis x line*=bottom,
    axis y line*=left,
    tick align=outside,
    tick style={black!60},
    xtick={-0.04,-0.03,-0.02,-0.01,0.00},
    scaled x ticks=false,
    xticklabel style={
        /pgf/number format/fixed,
        /pgf/number format/precision=2,
        font=\small
    },
    yticklabel style={font=\small},
    xmajorgrids=true,
    ymajorgrids=false,
    major grid style={dotted, gray!35},
    legend to name=fig:diversity-legend-single,
    legend columns=3,
    legend style={
        draw=none,
        fill=none,
        font=\small,
        /tikz/every even column/.append style={column sep=0.9em}
    },
    legend cell align=left,
    clip=false
]

\addplot[
    black!55,
    line width=0.65pt,
    forget plot
] coordinates {(0,0.45) (0,6.55)};

\addplot[
    only marks,
    mark=*,
    mark size=2.4pt,
    color=bgeblue,
    error bars/.cd,
    x dir=both,
    x explicit,
    error bar style={line width=0.65pt, bgeblue},
    error mark options={rotate=90, mark size=1.4pt, line width=0.65pt}
]
table[
    x=estimate,
    y=y,
    x error minus=errminus,
    x error plus=errplus
]{
estimate y errminus errplus
-0.021971 0.82 0.006418 0.006289
-0.014861 1.82 0.007016 0.006766
-0.018164 2.82 0.007066 0.007055
-0.024887 3.82 0.006709 0.006325
-0.020777 4.82 0.006444 0.006070
-0.020132 5.82 0.005834 0.005598
};
\addlegendentry{BGE-M3}

\addplot[
    only marks,
    mark=square*,
    mark size=2.2pt,
    color=e5green,
    error bars/.cd,
    x dir=both,
    x explicit,
    error bar style={line width=0.65pt, e5green},
    error mark options={rotate=90, mark size=1.4pt, line width=0.65pt}
]
table[
    x=estimate,
    y=y,
    x error minus=errminus,
    x error plus=errplus
]{
estimate y errminus errplus
-0.006064 1.00 0.002486 0.002512
-0.003031 2.00 0.001942 0.001943
-0.003906 3.00 0.001796 0.001964
-0.003229 4.00 0.002036 0.001997
-0.002389 5.00 0.002155 0.002106
-0.004725 6.00 0.001704 0.001754
};
\addlegendentry{E5-large}

\addplot[
    only marks,
    mark=triangle*,
    mark size=2.6pt,
    color=labseorange,
    error bars/.cd,
    x dir=both,
    x explicit,
    error bar style={line width=0.65pt, labseorange},
    error mark options={rotate=90, mark size=1.4pt, line width=0.65pt}
]
table[
    x=estimate,
    y=y,
    x error minus=errminus,
    x error plus=errplus
]{
estimate y errminus errplus
-0.016850 1.18 0.008858 0.009197
-0.015997 2.18 0.008333 0.008421
-0.017802 3.18 0.007140 0.007620
-0.023507 4.18 0.008127 0.007607
-0.022417 5.18 0.007620 0.007694
-0.026889 6.18 0.008696 0.008640
};
\addlegendentry{LaBSE}

\end{axis}
\end{tikzpicture}

\vspace{0.45em}

\begin{tikzpicture}
\begin{axis}[
    width=\linewidth,
    height=2.85cm,
    xmin=0.00,
    xmax=0.08,
    ymin=0.55,
    ymax=3.45,
    title={(B) Chinese--English gap},
    title style={font=\small},
    xlabel={Chinese $-$ English gap},
    xlabel style={font=\small},
    ytick={1,2,3},
    yticklabels={BGE-M3,E5-large,LaBSE},
    y dir=reverse,
    axis x line*=bottom,
    axis y line*=left,
    tick align=outside,
    tick style={black!60},
    xtick={0.00,0.02,0.04,0.06,0.08},
    scaled x ticks=false,
    xticklabel style={
        /pgf/number format/fixed,
        /pgf/number format/precision=2,
        font=\small
    },
    yticklabel style={font=\small},
    xmajorgrids=true,
    ymajorgrids=false,
    major grid style={dotted, gray!35},
    clip=false
]

\addplot[
    black!55,
    line width=0.65pt,
    forget plot
] coordinates {(0,0.55) (0,3.45)};

\addplot[
    only marks,
    mark=*,
    mark size=2.4pt,
    color=bgeblue,
    error bars/.cd,
    x dir=both,
    x explicit,
    error bar style={line width=0.65pt, bgeblue},
    error mark options={rotate=90, mark size=1.4pt, line width=0.65pt}
]
table[
    x=estimate,
    y=y,
    x error minus=errminus,
    x error plus=errplus
]{
estimate y errminus errplus
0.046861 0.90 0.006704 0.006259
};

\addplot[
    only marks,
    mark=o,
    mark size=2.4pt,
    color=bgeblue,
    fill=white,
    error bars/.cd,
    x dir=both,
    x explicit,
    error bar style={line width=0.65pt, bgeblue},
    error mark options={rotate=90, mark size=1.4pt, line width=0.65pt}
]
table[
    x=estimate,
    y=y,
    x error minus=errminus,
    x error plus=errplus
]{
estimate y errminus errplus
0.048700 1.10 0.007017 0.006742
};

\addplot[
    only marks,
    mark=square*,
    mark size=2.2pt,
    color=e5green,
    error bars/.cd,
    x dir=both,
    x explicit,
    error bar style={line width=0.65pt, e5green},
    error mark options={rotate=90, mark size=1.4pt, line width=0.65pt}
]
table[
    x=estimate,
    y=y,
    x error minus=errminus,
    x error plus=errplus
]{
estimate y errminus errplus
0.052546 1.90 0.001993 0.001985
};

\addplot[
    only marks,
    mark=square,
    mark size=2.2pt,
    color=e5green,
    fill=white,
    error bars/.cd,
    x dir=both,
    x explicit,
    error bar style={line width=0.65pt, e5green},
    error mark options={rotate=90, mark size=1.4pt, line width=0.65pt}
]
table[
    x=estimate,
    y=y,
    x error minus=errminus,
    x error plus=errplus
]{
estimate y errminus errplus
0.053886 2.10 0.002754 0.002712
};

\addplot[
    only marks,
    mark=triangle*,
    mark size=2.6pt,
    color=labseorange,
    error bars/.cd,
    x dir=both,
    x explicit,
    error bar style={line width=0.65pt, labseorange},
    error mark options={rotate=90, mark size=1.4pt, line width=0.65pt}
]
table[
    x=estimate,
    y=y,
    x error minus=errminus,
    x error plus=errplus
]{
estimate y errminus errplus
0.046580 2.90 0.009054 0.009191
};

\addplot[
    only marks,
    mark=triangle,
    mark size=2.6pt,
    color=labseorange,
    fill=white,
    error bars/.cd,
    x dir=both,
    x explicit,
    error bar style={line width=0.65pt, labseorange},
    error mark options={rotate=90, mark size=1.4pt, line width=0.65pt}
]
table[
    x=estimate,
    y=y,
    x error minus=errminus,
    x error plus=errplus
]{
estimate y errminus errplus
0.036541 3.10 0.009608 0.009545
};

\end{axis}
\end{tikzpicture}
\caption{Effect of the diversity-aware prompt.}
\label{fig:diversity-intervention}
\parbox{\linewidth}{\raggedright\footnotesize \textit{Note.} Negative values in Panel A indicate lower prior-argument similarity under the diversity prompt. In Panel B, filled markers show baseline estimates and open markers show diversity-prompt estimates; the Chinese--English gap remains positive.}
\end{figure}
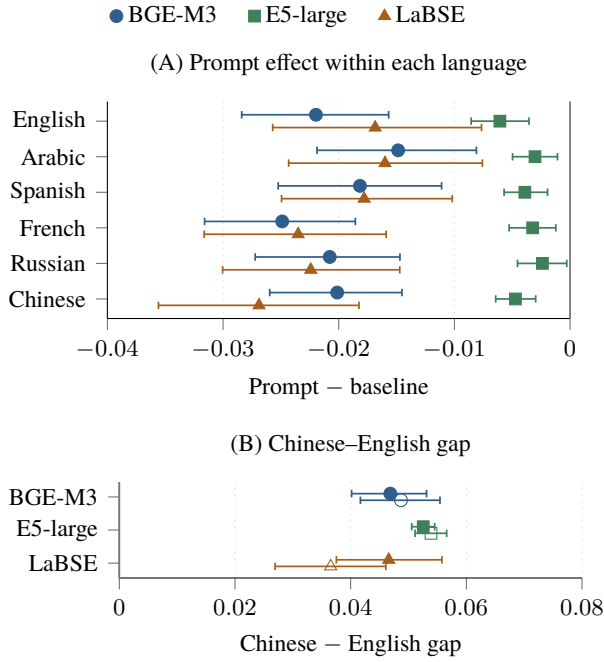

We test a simple mitigation: an extra instruction asking each speaker to introduce a new argument, example, mechanism, stakeholder impact, or tradeoff. The prompt works in the narrow sense that it lowers prior-argument similarity within languages. In Figure~\ref{fig:diversity-intervention}A, the prompt-minus-baseline effect is negative in all $18$ language--embedding cells. The reductions range from $-0.015$ to $-0.025$ under BGE-M3, $-0.002$ to $-0.006$ under multilingual-E5-large, and $-0.016$ to $-0.027$ under LaBSE. Output volume is preserved, so the reduction is not achieved by producing fewer extracted arguments.

The same prompt does not close the Chinese--English gap. In Figure~\ref{fig:diversity-intervention}B, the gap remains positive under the prompt for all three embeddings. It changes from $+0.047$ to $+0.049$ under BGE-M3, from $+0.053$ to $+0.054$ under multilingual-E5-large, and from $+0.047$ to $+0.037$ under LaBSE. None of the gap-reduction confidence intervals excludes zero: the prompt improves the overall level, but not the difference between Chinese and English. Exact estimates are in Appendix Tables~\ref{tab:app-diversity-within-lang} and~\ref{tab:app-diversity-zh-en-gap}.

This is not specific to the wording we chose. Under a six-condition ablation that also varies paraphrase, minimal, strong-novelty, and targeted-clash instructions, all 18 Chinese--English gaps stay positive with intervals excluding zero, and none of the 15 gap interactions is distinguishable from zero. Running on 10 motions, it is a lower-powered consistency check rather than a precise interaction estimate (Appendix~\ref{app:prompt-ablation}).

\subsection{Process similarity tracks judged outcomes but is not a stopping rule}
\label{sec:res-outcome-quality}

Across the 284 judged sessions, higher similarity is associated with lower judged quality: with motion fixed effects, a $+0.01$ increase predicts $-0.16$ points of argumentative development and $-0.10$ points of final argument quality, and all eight baseline language--dimension correlations are negative. The association does not transfer cleanly to interventions, however. The diversity prompt lowers similarity by an indistinguishable amount in English and Chinese ($-0.023$ vs. $-0.020$), yet its engagement and final-quality gains reach significance only in English. Equal movement in the process score therefore does not guarantee equal downstream benefit, and the score should be read as a process monitor rather than a standalone quality measure or a stopping rule (Appendix~\ref{app:outcome-quality}).

\subsection{Repeat generations support aggregate claims, not motion rankings}
\label{sec:res-repeat}

To check whether the findings depend on a single stochastic generation, we rerun a 10-motion subset with three independent generations per cell. The aggregate Chinese--English gap reappears in every baseline run under BGE-M3 ($+0.060$, $+0.043$, $+0.054$), and prior-argument similarity decreases for both English and Chinese in all three runs. Neither the contrast nor the prompt effect is a one-off generation artifact.

The same repeats show why we do not interpret individual motion rankings: motion-level gaps move substantially across generations. We therefore treat per-motion rankings as exploratory and base inference on aggregate language-level contrasts (Appendix~\ref{app:repeat-stability}).
\section{Discussion}
\label{sec:discussion}

In our matched pipeline, Chinese debates show higher prior-argument similarity than English, with manual calibration localizing the signal to a high-similarity tail enriched for substantive repetition. This brings an argumentation-theoretic view into LLM debate evaluation: if argumentation responds to a difference of opinion~\citep[p.~2]{vanEemeren2014handbook}, evaluation should ask not only what verdict a system reaches, but how argumentative ground changes.

\paragraph{From debate outcomes to argumentative state.} Prior work usually treats LLM debate as a procedure for improving an answer, score, or verdict~\citep{du2023improving,chan2023chateval,liang2024encouraging}. We extend this outcome-centered line of work by making the debate transcript itself measurable. \textit{Prior-argument similarity} asks whether a later turn adds claims, mechanisms, examples, stakeholder impacts, and tradeoffs already present, or returns to earlier ground in new wording. This metric does not replace answer accuracy or debate judging, and low similarity is not automatically better; rather, it provides an aggregate diagnostic of whether the interaction continues to expand the available argumentative material. The outcome judge supports this division of labor: higher similarity is moderately associated with lower judged development and final quality, so the diagnostic is not orthogonal to quality, yet comparable similarity reductions produce unequal downstream gains across languages, so it does not stand in for quality either.

\paragraph{From multilingual outputs to multilingual interaction.} Multilingual LLM evaluation has shown that language and culture affect model responses~\citep{shi2023language,arora2023probing,naous2024having,gao2025multilingualreasoning}. We extend this line of work from final outputs to interaction dynamics: even when the motion, model agents, roles, prompts, and turn positions are paired across languages, the debate process can still differ. The Chinese--English contrast should therefore be read as a model-language process effect under the tested pipeline, not as a claim about Chinese or English human argumentation. Methodologically, translated prompts or parallel tasks alone are insufficient for multilingual evaluation of interactive systems. If the object is a debate, the evaluation should also pair role assignments, turn order, prior context, and model identity, then inspect how content develops over time.

\paragraph{Candidate mechanisms.} Our checks narrow the space of explanations without closing it. Artifact diagnostics make the most proximate accounts unlikely, and a gap present for all four providers argues against a single-provider quirk. Translate-and-rescore then splits what remains: most of the raw gap travels with the surface language, while a smaller component survives translation into a common English embedding space. Which upstream factor produces that residual---pretraining composition, post-training, generation-side tokenization, provider settings, discourse conventions, or their interaction---is not identifiable in our design. This is a first decomposition, not a causal account.

\paragraph{Implications for AI-mediated deliberation.} Debate helps decision-makers compare reasons before reaching a conclusion~\citep{glazer2000debates}. Debate-style systems may help journalists test competing explanations, editors surface missing stakeholder perspectives, or policy analysts compare objections to regulatory proposals. In such settings, a fluent final summary can still be inadequate if interaction returns to earlier ground in polished wording rather than surfacing new considerations. Our findings suggest two checks: whether users in different languages receive comparable argumentative development, and whether mitigation improves the average process metric or also narrows language-conditioned gaps. The diversity-aware prompt lowered similarity overall but did not close the Chinese--English gap; a system-level improvement can therefore leave process differences intact. Its engagement and final-quality gains also appeared only in English. Multilingual mitigation needs both process-parity and outcome checks.
\section{Conclusion}
\label{sec:conclusion}

We studied multilingual LLM debate as a process of argumentative development, not only as a final-answer procedure. Using \textit{prior-argument similarity}, we find that Chinese is the only tested language with a sign-consistent positive gap relative to English across three multilingual embedding models. The gap persists across robustness checks, while manual calibration narrows the interpretation: the metric is not an item-level repetition classifier, but its high-similarity tail is enriched for substantive repetition. Translate-and-rescore further indicates that the measured contrast combines a substantial surface-language or representation component with a smaller surviving content-level component, and an independent outcome judge links higher similarity to moderately lower argumentative development and final quality.

The intervention result adds a second caution. A diversity-aware prompt lowers prior-argument similarity on average, but does not significantly narrow the Chinese--English gap, and the same contrast survives five alternative prompting strategies. Comparable process-score reductions can also yield unequal engagement and final-quality gains across languages. Multilingual debate systems should therefore be evaluated jointly for process development, downstream quality, and cross-language parity.

\section*{Limitations}
\label{sec:limitations}

This study has several scope and measurement limitations. The debate agents are API-served frontier systems: GPT-5.5, Claude Opus 4.7, DeepSeek-V3.2, and Mistral Large 3. These systems are relevant to user-facing multilingual generation, but they do not expose tokenizers, training corpora, post-training procedures, alignment data, or provider-side decoding implementations. The results therefore establish a pattern across the providers tested, not a claim about all open-weight models, smaller models, or future model generations.

Prior-argument similarity is a diagnostic rather than a direct human repetition label. The English--Chinese manual audit supports comparable extraction quality and shows that the high-similarity tail is enriched for substantive repetition, but item-level alignment with human repetition judgments is weak, and neither a cross-encoder nor a blinded LLM pair judge closes that gap. Manual calibration is also limited to English and Chinese, so the diagnostic meaning of the score in Arabic, Spanish, French, and Russian remains less directly validated.

The outcome-quality evidence is similarly scoped. Debate quality is scored by a single blinded LLM judge on English and Chinese sessions only. The judge is reliable on re-scoring and blinded to language condition and agent identity, but it uses a narrow range of integer values in some cells, may carry language-specific scale tendencies, and is not a substitute for human evaluation. We therefore report process--outcome associations as convergent validity rather than as a ranking of debate quality across languages.

The primary pipeline uses a single extractor, Llama-4-Scout-17B-16E-Instruct. Holding the extractor fixed makes cross-language processing consistent, but it may introduce language-specific boundary, omission, or field-quality biases. We report automatic diagnostics for all six languages, manually audit English and Chinese, and include a targeted English--Chinese second-extractor subset check. Stronger evidence would still require full multi-extractor robustness, extractor-prompt ablations, or manual extraction audits in more languages.

The setting is intentionally controlled: eight-turn parliamentary-style policy debates on WUDC motions, with fixed role assignments, turn schedules, and sentence-count constraints. Sentence-compliance diagnostics make a Chinese-specific template artifact unlikely, but the template may still shape generation and extraction. The language sample is also limited to six UN languages; Chinese is the only sign-consistent positive gap in this sample, not necessarily among world languages.

The study is descriptive rather than mechanistic. The translate-and-rescore probe separates a surface-language or representation component from a smaller content-level component, but it relies on machine translation and cannot attribute the residual to tokenization, pretraining data, post-training behavior, provider settings, discourse conventions, or their interaction. The intervention evidence is similarly scoped: one lightweight diversity prompt lowers the average metric without closing the Chinese--English gap, and the six-condition prompt ablation that reproduces this pattern covers only 10 motions, so it is a consistency check rather than a precise interaction estimate. Stronger novelty mechanisms or open-weight replications may behave differently.


\bibliography{custom}

\appendix
\section{Additional Result Tables}
\label{app:add-results}

This appendix reports exact estimates and auxiliary diagnostics summarized in Section~\ref{sec:results}. Positive language gaps indicate higher prior-argument similarity than English. Asterisks mark 95\% motion-cluster bootstrap confidence intervals that exclude zero. The core supporting results are in Tables~\ref{tab:app-lang-gaps}--\ref{tab:app-diversity-zh-en-gap} and Figures~\ref{fig:zh-en-robustness}--\ref{fig:turn-trajectory}. The remaining appendices report full agent-stratified estimates, prompt-strategy ablations, translation parameters and schemas, embedding calibration, translate-and-rescore, item-level instrument comparisons, multilingual examples, and the process--outcome evaluation.

\subsection{Full Six-Language Landscape}
\label{app:full-language-landscape}

Exact language gaps relative to English for all five non-English languages and all three embedding models are in Table~\ref{tab:app-lang-gaps}.

\begin{table*}[t]
\centering
\small
\begin{tabular}{lccccc}
\toprule
Embedding & Arabic & Spanish & French & Russian & Chinese \\
\midrule
BGE-M3                 & $-0.025^{*}$ & $-0.022^{*}$ & $-0.018^{*}$ & $-0.010^{*}$ & $\mathbf{+0.047^{*}}$ \\
multilingual-E5-large  & $+0.039^{*}$ & $+0.026^{*}$ & $+0.027^{*}$ & $+0.021^{*}$ & $\mathbf{+0.053^{*}}$ \\
LaBSE                  & $+0.029^{*}$ & $+0.058^{*}$ & $+0.065^{*}$ & $+0.083^{*}$ & $\mathbf{+0.047^{*}}$ \\
\bottomrule
\end{tabular}
\caption{Language gaps relative to English.}
\label{tab:app-lang-gaps}
\par\vspace{2pt}
\parbox{\linewidth}{\raggedright\footnotesize \textit{Note.} Gaps are in \texttt{mean\_arg\_max\_prev}; positive values indicate higher prior-argument similarity than English. Estimates are paired by motion and pooled across model agents. Asterisks mark 95\% motion-cluster bootstrap confidence intervals excluding zero.}
\end{table*}

\subsection{Pooled Chinese--English Gap}
\label{app:pooled-zh-en-gap}

The pooled Chinese--English gap and paired motion-cluster bootstrap confidence intervals for each embedding model are in Table~\ref{tab:app-pooled-zh-en}.

\begin{table}[!htbp]
\centering
\small
\begin{tabular}{@{}lcc@{}}
\toprule
Embedding & zh $-$ en & 95\% CI \\
\midrule
BGE-M3
  & $\mathbf{+0.047^{*}}$
  & $[+0.041,+0.053]$ \\
\shortstack[l]{multilingual-\\E5-large}
  & $\mathbf{+0.053^{*}}$
  & $[+0.050,+0.055]$ \\
LaBSE
  & $\mathbf{+0.047^{*}}$
  & $[+0.038,+0.056]$ \\
\bottomrule
\end{tabular}
\caption{Pooled Chinese--English gap.}
\label{tab:app-pooled-zh-en}
\par\vspace{2pt}
\parbox{\linewidth}{\raggedright\footnotesize \textit{Note.} Positive values indicate higher prior-argument similarity in Chinese than in English. Confidence intervals are paired motion-cluster bootstrap intervals; asterisks mark intervals excluding zero.}
\end{table}

\subsection{Model-Agent and Turn-Position Diagnostics}
\label{app:agent-turn-diagnostics}

\begin{figure}[!htbp]
\centering
\begin{tikzpicture}
\begin{axis}[
    width=0.78\linewidth,
    height=4.8cm,
    xmin=2,
    xmax=8,
    ymin=0.58,
    ymax=0.75,
    xlabel={Turn position},
    ylabel={Mean prior-argument similarity},
    xlabel style={font=\small},
    ylabel style={font=\small},
    xtick={2,3,4,5,6,7,8},
    ytick={0.58,0.60,0.62,0.64,0.66,0.68,0.70,0.72,0.74},
    scaled y ticks=false,
    yticklabel style={
        /pgf/number format/fixed,
        /pgf/number format/precision=2,
        font=\small
    },
    xticklabel style={font=\small},
    axis x line*=bottom,
    axis y line*=left,
    tick align=outside,
    tick style={black!60},
    xmajorgrids=false,
    ymajorgrids=true,
    major grid style={dotted, gray!35},
    legend columns=3,
    legend style={
        draw=none,
        fill=white,
        fill opacity=0.90,
        text opacity=1,
        font=\small,
        at={(0.50,1.10)},
        anchor=south,
        /tikz/every even column/.append style={column sep=1.4em}
    },
    legend cell align=left,
    clip=true
]

\addplot[
    color=contextgray,
    line width=0.75pt,
    mark=*,
    mark size=1.7pt,
    opacity=0.75
] coordinates {
    (2,0.590142)
    (3,0.621421)
    (4,0.630399)
    (5,0.635968)
    (6,0.648432)
    (7,0.672134)
    (8,0.661979)
};
\addlegendentry{Other non-English}

\addplot[
    color=contextgray,
    line width=0.75pt,
    mark=*,
    mark size=1.7pt,
    opacity=0.75,
    forget plot
] coordinates {
    (2,0.600477)
    (3,0.631058)
    (4,0.629532)
    (5,0.634218)
    (6,0.646587)
    (7,0.673225)
    (8,0.669159)
};

\addplot[
    color=contextgray,
    line width=0.75pt,
    mark=*,
    mark size=1.7pt,
    opacity=0.75,
    forget plot
] coordinates {
    (2,0.597741)
    (3,0.636130)
    (4,0.633648)
    (5,0.643290)
    (6,0.656615)
    (7,0.675018)
    (8,0.667363)
};

\addplot[
    color=contextgray,
    line width=0.75pt,
    mark=*,
    mark size=1.7pt,
    opacity=0.75,
    forget plot
] coordinates {
    (2,0.603669)
    (3,0.645717)
    (4,0.635424)
    (5,0.656693)
    (6,0.654097)
    (7,0.687307)
    (8,0.677538)
};

\addplot[
    color=bgeblue,
    line width=1.35pt,
    mark=square*,
    mark size=2.4pt
] coordinates {
    (2,0.616280)
    (3,0.647908)
    (4,0.654625)
    (5,0.660123)
    (6,0.672890)
    (7,0.695654)
    (8,0.684038)
};
\addlegendentry{English}

\addplot[
    color=accentred,
    line width=1.55pt,
    mark=*,
    mark size=2.6pt
] coordinates {
    (2,0.670473)
    (3,0.706970)
    (4,0.698865)
    (5,0.707861)
    (6,0.706846)
    (7,0.730706)
    (8,0.734872)
};
\addlegendentry{Chinese}

\end{axis}
\end{tikzpicture}
\caption{Turn-level prior-argument-similarity trajectories under BGE-M3.}
\label{fig:turn-trajectory}
\parbox{\linewidth}{\raggedright\footnotesize \textit{Note.} Prior-argument similarity increases as more prior arguments accumulate, but Chinese remains above English at every measured turn position. Gray lines show the other non-English languages for context.}
\end{figure}
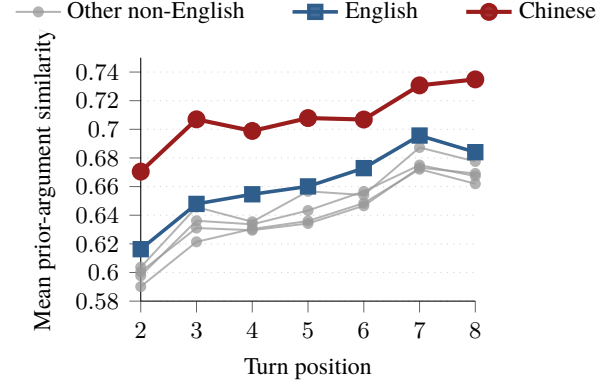

The pooled Chinese--English gap across embedding models and the corresponding model-agent gaps are plotted in Figure~\ref{fig:zh-en-robustness}. The BGE-M3 turn-level trajectory is plotted in Figure~\ref{fig:turn-trajectory}. These figures support the main-text claim that the focal contrast is not driven by a single provider, model agent, or late debate phase.

\begin{figure}[!htbp]
\centering

\pgfplotslegendfromname{fig:zh-en-legend-single}

\vspace{0.15em}

\begin{tikzpicture}
\begin{axis}[
    width=0.78\linewidth,
    height=2.65cm,
    xmin=0.00,
    xmax=0.08,
    ymin=0.55,
    ymax=3.45,
    title={(A) Pooled zh--en gap},
    title style={font=\small},
    xlabel={$\Delta$ prior-argument similarity},
    xlabel style={font=\small},
    ytick={1,2,3},
    yticklabels={BGE-M3,E5-large,LaBSE},
    y dir=reverse,
    axis x line*=bottom,
    axis y line*=left,
    tick align=outside,
    tick style={black!60},
    xtick={0.00,0.02,0.04,0.06,0.08},
    scaled x ticks=false,
    xticklabel style={
        /pgf/number format/fixed,
        /pgf/number format/precision=2,
        font=\small
    },
    yticklabel style={font=\small},
    xmajorgrids=true,
    ymajorgrids=false,
    major grid style={dotted, gray!35},
    legend to name=fig:zh-en-legend-single,
    legend columns=3,
    legend style={
        draw=none,
        fill=none,
        font=\small,
        /tikz/every even column/.append style={column sep=1.0em}
    },
    legend cell align=left,
    clip=false
]

\addplot[
    black!55,
    line width=0.6pt,
    forget plot
] coordinates {(0,0.55) (0,3.45)};

\addplot[
    only marks,
    mark=*,
    mark size=2.4pt,
    color=bgeblue,
    error bars/.cd,
    x dir=both,
    x explicit,
    error bar style={line width=0.65pt, bgeblue},
    error mark options={rotate=90, mark size=1.4pt, line width=0.65pt}
]
table[
    x=estimate,
    y=y,
    x error minus=errminus,
    x error plus=errplus
]{
estimate y errminus errplus
0.046861 1 0.006327 0.006244
};
\addlegendentry{BGE-M3}

\addplot[
    only marks,
    mark=square*,
    mark size=2.2pt,
    color=e5green,
    error bars/.cd,
    x dir=both,
    x explicit,
    error bar style={line width=0.65pt, e5green},
    error mark options={rotate=90, mark size=1.4pt, line width=0.65pt}
]
table[
    x=estimate,
    y=y,
    x error minus=errminus,
    x error plus=errplus
]{
estimate y errminus errplus
0.052546 2 0.002074 0.001978
};
\addlegendentry{multilingual-E5}

\addplot[
    only marks,
    mark=triangle*,
    mark size=2.6pt,
    color=labseorange,
    error bars/.cd,
    x dir=both,
    x explicit,
    error bar style={line width=0.65pt, labseorange},
    error mark options={rotate=90, mark size=1.4pt, line width=0.65pt}
]
table[
    x=estimate,
    y=y,
    x error minus=errminus,
    x error plus=errplus
]{
estimate y errminus errplus
0.046580 3 0.008831 0.009140
};
\addlegendentry{LaBSE}

\end{axis}
\end{tikzpicture}

\vspace{0.45em}

\begin{tikzpicture}
\begin{axis}[
    width=0.78\linewidth,
    height=3.85cm,
    xmin=0.00,
    xmax=0.08,
    ymin=0.45,
    ymax=4.55,
    title={(B) Model-agent gaps},
    title style={font=\small},
    xlabel={$\Delta$ prior-argument similarity},
    xlabel style={font=\small},
    ytick={1,2,3,4},
    yticklabels={GPT-5.5,Claude,DeepSeek,Mistral},
    y dir=reverse,
    axis x line*=bottom,
    axis y line*=left,
    tick align=outside,
    tick style={black!60},
    xtick={0.00,0.02,0.04,0.06,0.08},
    scaled x ticks=false,
    xticklabel style={
        /pgf/number format/fixed,
        /pgf/number format/precision=2,
        font=\small
    },
    yticklabel style={font=\small},
    xmajorgrids=true,
    ymajorgrids=false,
    major grid style={dotted, gray!35},
    clip=false
]

\addplot[
    black!55,
    line width=0.6pt,
    forget plot
] coordinates {(0,0.45) (0,4.55)};

\addplot[
    only marks,
    mark=*,
    mark size=2.4pt,
    color=bgeblue,
    error bars/.cd,
    x dir=both,
    x explicit,
    error bar style={line width=0.65pt, bgeblue},
    error mark options={rotate=90, mark size=1.4pt, line width=0.65pt}
]
table[
    x=estimate,
    y=y,
    x error minus=errminus,
    x error plus=errplus
]{
estimate y errminus errplus
0.033530 0.74 0.010829 0.010863
0.044515 1.74 0.010509 0.010585
0.048406 2.74 0.010708 0.010766
0.060993 3.74 0.009907 0.010244
};

\addplot[
    only marks,
    mark=square*,
    mark size=2.2pt,
    color=e5green,
    error bars/.cd,
    x dir=both,
    x explicit,
    error bar style={line width=0.65pt, e5green},
    error mark options={rotate=90, mark size=1.4pt, line width=0.65pt}
]
table[
    x=estimate,
    y=y,
    x error minus=errminus,
    x error plus=errplus
]{
estimate y errminus errplus
0.049232 1.00 0.003492 0.003343
0.051759 2.00 0.003870 0.003826
0.053001 3.00 0.003593 0.003404
0.056193 4.00 0.003179 0.003238
};

\addplot[
    only marks,
    mark=triangle*,
    mark size=2.6pt,
    color=labseorange,
    error bars/.cd,
    x dir=both,
    x explicit,
    error bar style={line width=0.65pt, labseorange},
    error mark options={rotate=90, mark size=1.4pt, line width=0.65pt}
]
table[
    x=estimate,
    y=y,
    x error minus=errminus,
    x error plus=errplus
]{
estimate y errminus errplus
0.028317 1.26 0.016065 0.015944
0.047423 2.26 0.015387 0.015739
0.050608 3.26 0.015838 0.015532
0.059973 4.26 0.014882 0.014905
};

\end{axis}
\end{tikzpicture}
\caption{Chinese--English gap across embeddings and model agents.}
\label{fig:zh-en-robustness}
\parbox{\linewidth}{\raggedright\footnotesize \textit{Note.} Panel A shows pooled gaps; Panel B shows model-agent gaps. Error bars denote 95\% motion-cluster bootstrap confidence intervals. All model--embedding cells are positive, indicating that the focal contrast is not driven by a single model agent or embedding space.}
\end{figure}
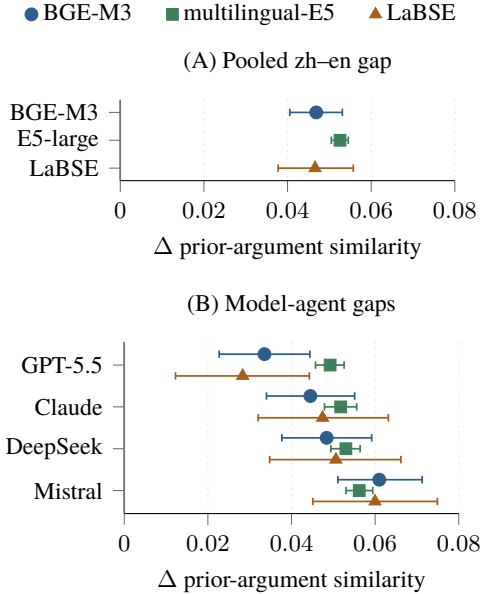

\subsection{Full Agent-Stratified Language Gaps}
\label{app:agent-language-gaps}

Table~\ref{tab:agent-language-gaps-full} reports every language--English contrast separately for the four debate agents and three embeddings. The focal Chinese--English gap is positive and statistically distinguishable from zero in all 12 agent--embedding cells, but its magnitude varies by agent. Four of the 60 total cells are not distinguishable from zero, all outside the focal contrast. Table~\ref{tab:agent-pairwise-focal} gives all 18 pairwise contrasts between agents with their bootstrap intervals: every point estimate is positive, but only six of the 18 intervals exclude zero, so the between-agent ordering is much less firmly established than the focal contrast itself.

\begin{table*}[t]
\centering
\small
\begin{tabular}{llrrrr}
\toprule
Embedding & Language $-$ English & GPT-5.5 & Claude & DeepSeek & Mistral \\
\midrule
BGE-M3 & Arabic  & $-0.037^{*}$ & $-0.029^{*}$ & $-0.029^{*}$ & $-0.003$ \\
       & Spanish & $-0.029^{*}$ & $-0.020^{*}$ & $-0.025^{*}$ & $-0.013^{*}$ \\
       & French  & $-0.021^{*}$ & $-0.019^{*}$ & $-0.020^{*}$ & $-0.010$ \\
       & Russian & $-0.018^{*}$ & $-0.013^{*}$ & $-0.012^{*}$ & $+0.006$ \\
       & Chinese & $+0.034^{*}$ & $+0.045^{*}$ & $+0.048^{*}$ & $+0.061^{*}$ \\
\midrule
multilingual-E5-large & Arabic  & $+0.036^{*}$ & $+0.038^{*}$ & $+0.037^{*}$ & $+0.043^{*}$ \\
       & Spanish & $+0.023^{*}$ & $+0.026^{*}$ & $+0.025^{*}$ & $+0.029^{*}$ \\
       & French  & $+0.024^{*}$ & $+0.028^{*}$ & $+0.027^{*}$ & $+0.030^{*}$ \\
       & Russian & $+0.018^{*}$ & $+0.018^{*}$ & $+0.021^{*}$ & $+0.025^{*}$ \\
       & Chinese & $+0.049^{*}$ & $+0.052^{*}$ & $+0.053^{*}$ & $+0.056^{*}$ \\
\midrule
LaBSE & Arabic  & $+0.014$ & $+0.024^{*}$ & $+0.027^{*}$ & $+0.050^{*}$ \\
       & Spanish & $+0.055^{*}$ & $+0.059^{*}$ & $+0.055^{*}$ & $+0.062^{*}$ \\
       & French  & $+0.061^{*}$ & $+0.065^{*}$ & $+0.061^{*}$ & $+0.074^{*}$ \\
       & Russian & $+0.086^{*}$ & $+0.080^{*}$ & $+0.075^{*}$ & $+0.092^{*}$ \\
       & Chinese & $+0.028^{*}$ & $+0.047^{*}$ & $+0.051^{*}$ & $+0.060^{*}$ \\
\bottomrule
\end{tabular}
\caption{Agent-stratified language--English gaps in prior-argument similarity. Asterisks indicate that the 95\% paired motion-cluster bootstrap CI excludes zero.}
\label{tab:agent-language-gaps-full}
\par\vspace{2pt}
\parbox{\linewidth}{\raggedright\footnotesize \textit{Note.} Each cell averages within motion and agent before differencing the target language from English; uncertainty uses 5{,}000 motion-cluster bootstrap resamples (seed 42). All cells contain 71 paired motions except Russian--Mistral ($n=70$). In M024-ru, Mistral occupied turns 1 and 5; turn 1 is excluded by construction and turn 5 was one of the two documented extraction parse failures, leaving no scoreable Mistral turn in that session.}
\end{table*}

\begin{table*}[t]
\centering
\small
\setlength{\tabcolsep}{3pt}
\begin{tabular}{lccc}
\toprule
Agent contrast & BGE-M3 & multilingual-E5-large & LaBSE \\
\midrule
Claude $-$ GPT-5.5   & $+0.0110$ $[-0.0026,+0.0248]$ & $+0.0025$ $[-0.0024,+0.0072]$ & $+0.0191$ $[-0.0020,+0.0401]$ \\
DeepSeek $-$ GPT-5.5 & $+0.0149$ $[+0.0003,+0.0294]^{*}$ & $+0.0038$ $[-0.0014,+0.0092]$ & $+0.0223$ $[+0.0005,+0.0447]^{*}$ \\
Mistral $-$ GPT-5.5  & $+0.0275$ $[+0.0157,+0.0396]^{*}$ & $+0.0070$ $[+0.0032,+0.0108]^{*}$ & $+0.0316$ $[+0.0141,+0.0483]^{*}$ \\
DeepSeek $-$ Claude  & $+0.0039$ $[-0.0101,+0.0183]$ & $+0.0012$ $[-0.0034,+0.0063]$ & $+0.0032$ $[-0.0163,+0.0234]$ \\
Mistral $-$ Claude   & $+0.0165$ $[+0.0022,+0.0310]^{*}$ & $+0.0044$ $[-0.0006,+0.0095]$ & $+0.0125$ $[-0.0100,+0.0346]$ \\
Mistral $-$ DeepSeek & $+0.0126$ $[-0.0002,+0.0259]$ & $+0.0032$ $[-0.0012,+0.0075]$ & $+0.0094$ $[-0.0116,+0.0303]$ \\
\bottomrule
\end{tabular}
\caption{Complete pairwise contrasts in the focal Chinese--English gap, with paired motion-cluster bootstrap confidence intervals. Positive values mean that the first agent has the larger gap.}
\label{tab:agent-pairwise-focal}
\par\vspace{2pt}
\parbox{\linewidth}{\raggedright\footnotesize \textit{Note.} Each contrast differences the agent-specific Chinese--English gaps of Table~\ref{tab:agent-language-gaps-full} within motion, then resamples the 71 motions 5{,}000 times with seed 42; intervals are percentile intervals. Asterisks mark the six of 18 intervals that exclude zero. Every point estimate is positive, so no agent reverses the focal sign; the individually distinguishable contrasts are those separating Mistral or DeepSeek from GPT-5.5, and Mistral from Claude. The design cannot identify a causal model-family or size effect because each closed provider contributes one model and comparable parameter counts are unavailable.}
\end{table*}

\subsection{Translation-QA Sensitivity}
\label{app:translation-sensitivity}

The Chinese--English gap after excluding increasingly conservative sets of translation-flagged motions is in Table~\ref{tab:translation-sensitivity}. The estimates remain stable under all exclusions.

\begin{table*}[!htbp]
\centering
\small
\begin{tabularx}{\textwidth}{@{}l c *{3}{>{\centering\arraybackslash}X}@{}}
\toprule
Subset & $n$ motions & BGE-M3 & mE5-large & LaBSE \\
\midrule
Full set & 71
& \makecell[c]{$+0.0469$\\$[+0.0406,+0.0530]$}
& \makecell[c]{$+0.0525$\\$[+0.0504,+0.0546]$}
& \makecell[c]{$+0.0466$\\$[+0.0376,+0.0555]$} \\

Excl. M056 & 70
& \makecell[c]{$+0.0477$\\$[+0.0414,+0.0542]$}
& \makecell[c]{$+0.0525$\\$[+0.0504,+0.0545]$}
& \makecell[c]{$+0.0468$\\$[+0.0377,+0.0560]$} \\

Excl. zh-flagged & 64
& \makecell[c]{$+0.0466$\\$[+0.0396,+0.0531]$}
& \makecell[c]{$+0.0522$\\$[+0.0499,+0.0543]$}
& \makecell[c]{$+0.0457$\\$[+0.0364,+0.0550]$} \\

Excl. all flagged & 51
& \makecell[c]{$+0.0454$\\$[+0.0377,+0.0527]$}
& \makecell[c]{$+0.0520$\\$[+0.0497,+0.0544]$}
& \makecell[c]{$+0.0472$\\$[+0.0378,+0.0570]$} \\
\bottomrule
\end{tabularx}
\caption{Translation-QA sensitivity.}
\label{tab:translation-sensitivity}
\par\vspace{2pt}
\parbox{\linewidth}{\raggedright\footnotesize \textit{Note.} Rows re-bootstrap the baseline Chinese--English gap after excluding increasingly conservative sets of flagged motions. The gap remains stable under all exclusions.}
\end{table*}

\subsection{Diversity-Prompt Effects}
\label{app:diversity-results-tables}

Within-language changes from the diversity prompt are in Table~\ref{tab:app-diversity-within-lang}. Gap-reduction estimates for the same intervention are in Table~\ref{tab:app-diversity-zh-en-gap}.

\begin{table*}[!htbp]
\centering
\small
\begin{tabular}{lcccccc}
\toprule
Embedding & English & Arabic & Spanish & French & Russian & Chinese \\
\midrule
BGE-M3                 & $-0.022^{*}$ & $-0.015^{*}$ & $-0.018^{*}$ & $-0.025^{*}$ & $-0.021^{*}$ & $-0.020^{*}$ \\
multilingual-E5-large  & $-0.006^{*}$ & $-0.003^{*}$ & $-0.004^{*}$ & $-0.003^{*}$ & $-0.002^{*}$ & $-0.005^{*}$ \\
LaBSE                  & $-0.017^{*}$ & $-0.016^{*}$ & $-0.018^{*}$ & $-0.024^{*}$ & $-0.022^{*}$ & $-0.027^{*}$ \\
\bottomrule
\end{tabular}
\caption{Within-language effect of the diversity prompt.}
\label{tab:app-diversity-within-lang}
\par\vspace{2pt}
\parbox{\linewidth}{\raggedright\footnotesize \textit{Note.} Values are diversity minus baseline; negative values indicate lower prior-argument similarity under the diversity prompt. Asterisks mark 95\% motion-cluster bootstrap confidence intervals excluding zero.}
\end{table*}

\begin{table*}[!htbp]
\centering
\small
\begin{tabular}{lcccc}
\toprule
Embedding & Baseline zh $-$ en & Diversity zh $-$ en & Gap reduction & 95\% CI \\
\midrule
BGE-M3                 & $+0.047$ & $+0.049$ & $-0.002$ & $[-0.010,\ +0.007]$ \\
multilingual-E5-large  & $+0.053$ & $+0.054$ & $-0.001$ & $[-0.004,\ +0.002]$ \\
LaBSE                  & $+0.047$ & $+0.037$ & $+0.010$ & $[-0.003,\ +0.023]$ \\
\bottomrule
\end{tabular}
\caption{Effect of the diversity prompt on the Chinese--English gap.}
\label{tab:app-diversity-zh-en-gap}
\par\vspace{2pt}
\parbox{\linewidth}{\raggedright\footnotesize \textit{Note.} Positive gap-reduction values indicate narrowing of the zh$-$en gap. No gap-reduction confidence interval excludes zero.}
\end{table*}

\subsection{Six-Condition Prompt-Strategy Ablation}
\label{app:prompt-ablation}

This ablation asks whether the focal contrast depends on the particular baseline and diversity wordings. It uses the pre-defined 10-motion English--Chinese repeat-stability subset, three independent generations per motion--language--condition cell, the locked role-to-model mappings, the same eight-turn structure, provider-default decoding, and the original extraction and scoring pipeline. Baseline and diversity-aware cells reuse the locked main generation and two repeat generations. Four new conditions---P1 paraphrase, P2 minimal, P3 strong-novelty, and P4 targeted-clash---add 240 sessions. Inference averages the four agents within motion--language--generation, then averages the three generations and bootstraps motions 5{,}000 times with seed 42. Table~\ref{tab:prompt-ablation-gaps} reports the gap under each condition and Table~\ref{tab:prompt-ablation-interactions} the condition-minus-baseline change.

\begin{table*}[t]
\centering
\small
\begin{tabular}{lccc}
\toprule
Condition & BGE-M3 & multilingual-E5-large & LaBSE \\
\midrule
Baseline          & $+0.0522$ $[+0.0422,+0.0617]$ & $+0.0563$ $[+0.0526,+0.0601]$ & $+0.0537$ $[+0.0364,+0.0726]$ \\
Diversity-aware   & $+0.0374$ $[+0.0235,+0.0520]$ & $+0.0526$ $[+0.0451,+0.0593]$ & $+0.0350$ $[+0.0088,+0.0626]$ \\
P1 paraphrase     & $+0.0499$ $[+0.0388,+0.0612]$ & $+0.0539$ $[+0.0501,+0.0579]$ & $+0.0485$ $[+0.0268,+0.0708]$ \\
P2 minimal        & $+0.0447$ $[+0.0329,+0.0557]$ & $+0.0539$ $[+0.0515,+0.0561]$ & $+0.0460$ $[+0.0367,+0.0562]$ \\
P3 strong-novelty & $+0.0537$ $[+0.0432,+0.0639]$ & $+0.0562$ $[+0.0498,+0.0614]$ & $+0.0420$ $[+0.0167,+0.0653]$ \\
P4 targeted-clash & $+0.0538$ $[+0.0426,+0.0652]$ & $+0.0548$ $[+0.0500,+0.0595]$ & $+0.0365$ $[+0.0169,+0.0566]$ \\
\bottomrule
\end{tabular}
\caption{Chinese--English gap under six prompt conditions. Every 95\% motion-cluster bootstrap CI excludes zero.}
\label{tab:prompt-ablation-gaps}
\end{table*}

\begin{table*}[t]
\centering
\small
\begin{tabular}{lrrr}
\toprule
Condition $-$ baseline & BGE-M3 & multilingual-E5-large & LaBSE \\
\midrule
Diversity-aware   & $-0.0148$ & $-0.0037$ & $-0.0187$ \\
P1 paraphrase     & $-0.0023$ & $-0.0024$ & $-0.0052$ \\
P2 minimal        & $-0.0075$ & $-0.0024$ & $-0.0077$ \\
P3 strong-novelty & $+0.0015$ & $-0.0001$ & $-0.0117$ \\
P4 targeted-clash & $+0.0016$ & $-0.0015$ & $-0.0172$ \\
\bottomrule
\end{tabular}
\caption{Change in the Chinese--English gap relative to baseline. None of the 15 interaction CIs excludes zero.}
\label{tab:prompt-ablation-interactions}
\par\vspace{2pt}
\parbox{\linewidth}{\raggedright\footnotesize \textit{Note.} Values are paired condition-minus-baseline changes in the language gap. All 15 percentile CIs contain zero. The largest absolute movement is the diversity-aware condition under LaBSE, $-0.0187$ $[-0.0505,+0.0123]$; under the primary BGE-M3 embedding it is $-0.0147$ $[-0.0335,+0.0028]$. With only 10 motions, these are consistency checks rather than precise interaction estimates.}
\end{table*}

Prompt strategy does change absolute similarity, even though it does not move the gap. P3 strong-novelty lowers BGE-M3 similarity by $-0.029$ in English and $-0.027$ in Chinese, and lowers LaBSE similarity by $-0.023$ and $-0.035$, respectively; all four CIs exclude zero. P4 targeted-clash lowers similarity in both languages under BGE-M3, while the diversity-aware rule has its clearest effects in Chinese. Format behavior remains within the normal range across conditions: 3.7--4.0 extracted arguments per turn and 90.0--96.7\% sentence-count compliance. Prompting therefore moves the level of recycling, but no tested strategy eliminates or reverses the focal gap.

\section{Experimental Materials and Data Audit}
\label{app:method-details}

\subsection{Debate Template and Yoked Role Mapping}
\label{app:debate-template}

Each debate session contains eight turns in a fixed order: two opening statements, two rebuttals, two cross-examination-style turns, and two closing turns. Four model agents participate in each session by occupying the four debate roles: Pro First Speaker, Con First Speaker, Pro Second Speaker, and Con Second Speaker.

The four provider-hosted model agents are OpenAI GPT-5.5~\citep{openai2026gpt55}, Anthropic Claude Opus 4.7~\citep{anthropic2026opus47}, DeepSeek-V3.2~\citep{deepseek2025v32}, and Mistral Large 3~\citep{mistral2025large3}. The DeepSeek agent was accessed as DeepSeek-Chat via the \texttt{deepseek-chat} endpoint, and the Mistral agent was accessed as Mistral Large via the \texttt{mistral-large-latest} endpoint. Both endpoint mappings were recorded on 2026-04-30, when they corresponded to DeepSeek-V3.2 and Mistral Large 3, respectively. We refer to these agents by shortened labels in some figures and tables.

For each motion, the assignment of model agents to debate roles is fixed deterministically across all six language conditions. This yokes model identity and role position to motion, so paired language contrasts do not re-sample which model acts as which speaker. For sessions sharing the same underlying motion, the main between-session variable across language conditions is therefore the language of the debate.

\begin{figure*}[!htbp]
\centering
\small
\resizebox{\linewidth}{!}{%
\begin{tikzpicture}[
    x=1cm,
    y=1cm,
    font=\small,
    inputbox/.style={
        draw=black!65,
        fill=black!3,
        rounded corners=1pt,
        align=center,
        inner sep=4pt,
        minimum width=3.25cm,
        minimum height=0.72cm
    },
    turnbox/.style={
        draw=black!70,
        rounded corners=1pt,
        align=center,
        inner sep=2.2pt,
        minimum width=1.72cm,
        minimum height=0.86cm
    },
    proturn/.style={turnbox, fill=bgeblue!8},
    conturn/.style={turnbox, fill=accentred!8},
    processbox/.style={
        draw=black!65,
        fill=black!3,
        rounded corners=1pt,
        align=center,
        inner sep=4pt,
        minimum width=3.4cm,
        minimum height=0.72cm
    },
    flow/.style={->, line width=0.55pt, black!70},
    contextflow/.style={->, dashed, line width=0.45pt, black!55},
    band/.style={draw=black!35, fill=black!2, rounded corners=1pt}
]

\node[inputbox] (motion) at (0,3.35) {Motion $m$\\Target language $\ell$};
\node[inputbox] (roles) at (5.05,3.35) {Yoked model--role map\\Agent A--D $\rightarrow$ four roles};
\node[inputbox] (template) at (10.10,3.35) {Turn prompt template\\role, side, phase, context};

\draw[flow] (motion.east) -- (roles.west);
\draw[flow] (roles.east) -- (template.west);

\node[proturn] (t1) at (0,1.70) {\textbf{T1}\\Opening\\Pro First};
\node[conturn] (t2) at (1.95,1.70) {\textbf{T2}\\Opening\\Con First};
\node[proturn] (t3) at (3.90,1.70) {\textbf{T3}\\Rebuttal\\Pro First};
\node[conturn] (t4) at (5.85,1.70) {\textbf{T4}\\Rebuttal\\Con First};
\node[proturn] (t5) at (7.80,1.70) {\textbf{T5}\\Cross-exam.\\Pro Second};
\node[conturn] (t6) at (9.75,1.70) {\textbf{T6}\\Cross-exam.\\Con Second};
\node[proturn] (t7) at (11.70,1.70) {\textbf{T7}\\Closing\\Pro Second};
\node[conturn] (t8) at (13.65,1.70) {\textbf{T8}\\Closing\\Con Second};

\draw[flow] (template.south) -- ++(0,-0.42) -| (t1.north);
\draw[flow] (t1.east) -- (t2.west);
\draw[flow] (t2.east) -- (t3.west);
\draw[flow] (t3.east) -- (t4.west);
\draw[flow] (t4.east) -- (t5.west);
\draw[flow] (t5.east) -- (t6.west);
\draw[flow] (t6.east) -- (t7.west);
\draw[flow] (t7.east) -- (t8.west);

\draw[band] (-0.86,0.36) rectangle (14.51,0.93);
\node[align=center, font=\footnotesize] at (6.82,0.65)
{For turn $t>1$: \texttt{previous\_speech}$=T_{t-1}$ and \texttt{debate\_history}$=(T_1,\ldots,T_{t-1})$};
\draw[contextflow] (t1.south) -- ++(0,-0.25);
\draw[contextflow] (t2.south) -- ++(0,-0.25);
\draw[contextflow] (t3.south) -- ++(0,-0.25);
\draw[contextflow] (t4.south) -- ++(0,-0.25);
\draw[contextflow] (t5.south) -- ++(0,-0.25);
\draw[contextflow] (t6.south) -- ++(0,-0.25);
\draw[contextflow] (t7.south) -- ++(0,-0.25);

\node[processbox] (transcript) at (4.10,-0.68) {Complete debate transcript\\eight constrained turns};
\node[processbox] (extract) at (9.25,-0.68) {Argument extraction\\1--4 units per turn};
\node[processbox] (prior) at (13.65,-0.68) {Prior-argument pool\\compare later units to earlier units};

\draw[flow] (t8.south) -- ++(0,-1.24) -| (transcript.north);
\draw[flow] (transcript.east) -- (extract.west);
\draw[flow] (extract.east) -- (prior.west);

\end{tikzpicture}
}
\caption{Debate generation protocol.}
\label{fig:debate-process}
\parbox{\linewidth}{\raggedright\footnotesize \textit{Note.} Each motion is debated in each target language using the same role order and a deterministic model-to-role assignment. At every turn after the first, the current speaker receives the immediately preceding speech and the accumulated debate history before generating the next response. The completed transcript is then passed to argument extraction and prior-argument-similarity scoring.}
\end{figure*}
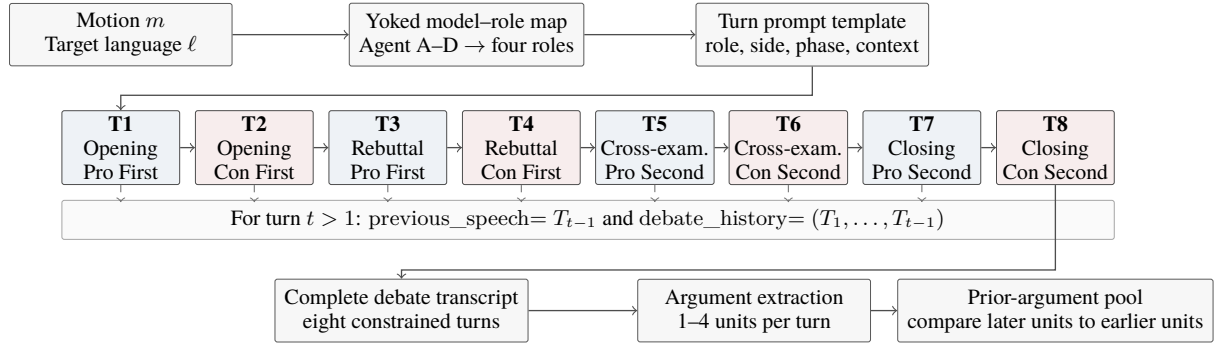

Figure~\ref{fig:debate-process} summarizes the session-level generation protocol. The corresponding eight-turn template is listed in Table~\ref{tab:debate-template}.

\begin{table}[!htbp]
\centering
\small
\begin{tabular}{cll}
\toprule
Turn & Phase & Role \\
\midrule
1 & Opening & Pro First Speaker \\
2 & Opening & Con First Speaker \\
3 & Rebuttal & Pro First Speaker \\
4 & Rebuttal & Con First Speaker \\
5 & Cross-exam. & Pro Second Speaker \\
6 & Cross-exam. & Con Second Speaker \\
7 & Closing & Pro Second Speaker \\
8 & Closing & Con Second Speaker \\
\bottomrule
\end{tabular}
\caption{Eight-turn debate template.}
\label{tab:debate-template}
\end{table}

\subsection{Translation Quality Review}
\label{app:translation-qa}

The 71 English debate motions were translated into Chinese, Spanish, French, Russian, and Arabic using Google Translate with default settings, yielding 355 non-English motion--language cells (and 426 cells including the English originals). The translated set was locked on 2026-04-29. Each non-English cell then passed through two automated QA stages followed by manual triage; Table~\ref{tab:translation-settings} lists every setting.

\begin{table*}[t]
\centering
\small
\begin{tabular}{lp{0.70\textwidth}}
\toprule
Component & Setting \\
\midrule
Initial translation & Google Translate, default settings; English to zh/es/fr/ru/ar; set locked 2026-04-29 \\
Coverage & 71 motions $\times$ five target languages $=355$ non-English cells \\
Back-translation & \texttt{deepseek-chat}; target text only; JSON-only; \texttt{max\_tokens=400}; provider-default temperature and top-$p$ \\
Adequacy judge & Llama-4-Scout-17B-16E-Instruct served through vLLM \\
Judge decoding & temperature $=0$; \texttt{max\_tokens=700} \\
Judge fields & 1--5 adequacy; preservation of actor, policy action, stance/burden, scope, and key terms; structured issue type \\
Flag rule & score $\leq3$, any preservation field false, or issue type other than \texttt{none} \\
Manual triage & A: meaning-level/must-fix; B: nuance; C: safe or back-translation artifact \\
Final audit & 43 cells initially flagged; 36/355 residual flags across 20 motions after correction and review; seven motions had Chinese-specific flags \\
\bottomrule
\end{tabular}
\caption{Translation and QA settings.}
\label{tab:translation-settings}
\end{table*}

No locked translation received an adequacy score of 3 or below; flags arose from the preservation fields or structured issue type. Manual review corrected meaning-changing cases and retained documented nuance or back-translation artifacts, including the reviewed borderline M056 Chinese motion. We do not remove residual flags from the primary analysis because post-outcome deletion could add researcher degrees of freedom. Instead, the sensitivity analysis excludes, in turn, M056, all seven motions with Chinese-specific flags, and all 20 motions with any residual flag.

\paragraph{Back-translation output schema.}
The back-translator saw only the target-language motion and returned one JSON object. For reproducibility, the operational content and normalized schema were:
\begin{PromptBlock}
Translate the following debate motion into English as literally and completely as possible.
Do not evaluate or explain it. Return JSON only.

Target-language motion: {target_motion}

{"back_translation_en": "..."}
\end{PromptBlock}

\paragraph{Adequacy-judge output schema.}
The adequacy judge compared the English original, target-language translation, and English back-translation and returned the following structured fields:
\begin{PromptBlock}
Compare the original motion, target-language translation, and back-translation.
Judge whether the target preserves the original debate burden and meaning.
Return JSON only.

{"adequacy_score": 1-5,
 "preserves_actor": true|false,
 "preserves_policy_action": true|false,
 "preserves_stance_or_burden": true|false,
 "preserves_scope": true|false,
 "preserves_key_terms": true|false,
 "issue_type": "none|actor|policy_action|stance_burden|scope|key_terms|other",
 "brief_reason": "..."}
\end{PromptBlock}
The displayed wording normalizes wrapper text for readability while preserving the operational inputs, decision fields, and flagging logic used in the pipeline.

\subsection{Session Counts and Execution Completeness}
\label{app:session-counts}

The baseline experiment contains 426 complete sessions, 3{,}408 turns, and 13{,}118 extracted argument units. The diversity-prompt experiment contains another 426 complete sessions, 3{,}408 turns, and 13{,}223 extracted argument units. The repeat-stability experiment adds 80 sessions, 640 turns, and 2{,}497 extracted argument units. Across all phases, the experiment contains 932 sessions, 7{,}456 turns, and 28{,}838 extracted argument units.

\subsection{Dataset Structure and Representative Record}
\label{app:dataset-example}
The primary baseline data form a complete 71-motion $\times$ six-language grid. Every session contains the same eight role-labeled turn positions, and each motion carries one fixed role-to-model mapping across its six language versions. The session table stores motion identifier and text, language, condition, role mapping, and execution status; the turn table stores session identifier, turn index and label, side, role, model, raw text, and prompt metadata; the argument table stores the extracted claim, warrant, evidence or example, type, and links to its turn and nearest prior argument.

\begin{table*}[t]
\centering
\small
\begin{tabular}{ll}
\toprule
Field & Example value \\
\midrule
Motion/session & M037, English, baseline \\
Unit & current argument and its nearest prior argument \\
Prior claim & ``art can be assessed objectively'' \\
Current claim & ``art can be evaluated objectively'' \\
BGE-M3 max-prior similarity & $0.997$ \\
Interpretation & near-duplicate substantive claim with different surface wording \\
\bottomrule
\end{tabular}
\caption{Representative processed high-similarity record from an English baseline session.}
\label{tab:dataset-record-example}
\end{table*}

API completion was effectively complete in the final locked outputs. The baseline and diversity phases each completed 3{,}408 of 3{,}408 expected turns, and the repeat phase completed 640 of 640 expected turns. Transient retry cases were retained only after successful completion and normal extraction.

\subsection{Operational Deviations}
\label{app:operational-deviations}

The diversity-prompt experiment was expanded from the originally planned English--Chinese subset to all six languages. The locked role mappings and language orders from the earlier session grid were reused unchanged; only the language coverage of the diversity prompt changed.

A mid-run provider credit-balance failure occurred during the first diversity-prompt launch. Partial sessions from that failed launch were excluded, and the run was resumed by session identifier after service restoration. The final diversity-prompt dataset contains the full expected 426 sessions.

\subsection{Data Artifacts}
\label{app:data-artifacts}

The primary processed data consist of per-session, per-turn, and per-argument metrics for the baseline, diversity-prompt, and repeat-stability phases. The raw debate sessions, turns, prompts, API-error logs, translation flags, and repeat-stability tables are retained in the experiment archive. The main analysis relies on locked processed outputs. The post-experiment diagnostics added for measurement validation are separately scoped: they re-extract a 30-motion English--Chinese subset with a second extractor and rescore existing English--Chinese max-prior pairs with a cross-encoder reranker; they do not regenerate debate transcripts. The archive additionally retains the 240 newly generated prompt-ablation sessions, translated and round-trip transcript manifests for translate-and-rescore, per-pair cross-encoder and LLM-judge scores, and blinded outcome-judge scores. Apart from the four new prompt conditions, all of these analyses read the locked debate archive without regenerating the original baseline or diversity sessions.

\section{Prompt Templates}
\label{app:prompts}

This appendix reports the prompt templates used to generate debate sessions. Appendix~\ref{app:baseline-prompt} gives the baseline turn-level prompt, and Appendix~\ref{app:diversity-prompt} gives the additional diversity-aware rule used in the intervention condition.

\paragraph{Provenance.}
The baseline prompt was authored by the authors and held fixed across all sessions, except for turn-specific placeholders such as the motion, target language, role, turn label, previous speech, and debate history. The diversity-aware rule was first written in English and then rendered in the other five target languages. The translated versions were checked as part of the multilingual material review described in Appendix~\ref{app:translation-qa}. Across languages, the rule preserves the same five novelty targets: argument, example, causal mechanism, stakeholder impact, and tradeoff.

\subsection{Baseline Prompt Template}
\label{app:baseline-prompt}

The template below is rendered once per turn. Placeholders in \texttt{\{...\}} are filled at inference time: \texttt{\{motion\_text\}} is the motion in the target language; \texttt{\{language\_name\}} is the natural-language name of the target language; \texttt{\{role\}}, \texttt{\{side\}}, and \texttt{\{turn\_label\}} identify the current speaker and turn; \texttt{\{previous\_speech\}} is the immediately preceding speech, or \texttt{(none --- this is the first turn)} for the first turn; and \texttt{\{debate\_history\}} concatenates all prior turns, or \texttt{(empty --- this is the first turn)} when no prior turn exists.

\begin{PromptBlock}
You are participating in a structured debate about:

"{motion_text}"

You must write only in {language_name}.

Your role is: {role}.
Your side is: {side}.
Current turn: {turn_label}.

Rules:
- Defend your assigned side.
- Respond to the previous speaker when previous context exists.
- Use clear reasoning.
- Do not announce your role.
- Do not greet the audience.
- Do not use bullet points, numbered lists, headings, or markdown.
- Do not fabricate statistics.
- Keep the response concise.

Length:
- Opening, rebuttal, and cross-examination turns: exactly 5 sentences.
- Closing turns: exactly 4 sentences.
- Each sentence should be under 50 words.

Previous speaker:
{previous_speech}

Debate history:
{debate_history}

Now produce your turn.
\end{PromptBlock}

\subsection{Diversity-Aware Intervention Rule}
\label{app:diversity-prompt}

Under the diversity-aware condition, the following rule is inserted into the baseline template immediately after the rules block and before the length block. The inserted rule is written in the target language of the session.

\paragraph{English}
\begin{quote}\small
Additional diversity rule:\\
In this turn, introduce at least one argument, example, causal mechanism, stakeholder impact, or tradeoff that has not appeared earlier in the debate. Do not merely restate your teammate's framing or the previous speaker's framing. If you refer to an earlier point, extend it with a clearly new reason, example, mechanism, or implication.
\end{quote}

\paragraph{Chinese}
{\cjkfont
\begin{quote}\small
额外的论点多样性规则：\\
在本轮发言中，请提出至少一个此前辩论中\\
没有出现过的论点、例子、因果机制、\\
利益相关方影响或权衡。\\
不要只是重复队友或上一位发言者的表述。\\
如果你引用此前观点，必须用一个清楚的\\
新理由、新例子、新机制或新影响\\
来扩展它。
\end{quote}
}

\paragraph{Spanish}
\begin{quote}\small
Regla adicional de diversidad de argumentos:\\
En este turno, introduzca al menos un argumento, ejemplo, mecanismo causal, impacto sobre las partes interesadas o compensación que no haya aparecido antes en el debate. No se limite a repetir el planteamiento de su compañero ni el del orador anterior. Si hace referencia a un punto anterior, amplíelo con una razón, un ejemplo, un mecanismo o una implicación claramente nuevos.
\end{quote}

\paragraph{French}
\begin{quote}\small
Règle supplémentaire de diversité des arguments :\\
Dans ce tour, introduisez au moins un argument, un exemple, un mécanisme causal, un impact sur les parties prenantes ou un compromis qui ne soit pas déjà apparu plus tôt dans le débat. Ne vous contentez pas de reformuler la formulation de votre coéquipier ou de l'orateur précédent. Si vous faites référence à un point antérieur, prolongez-le avec une raison, un exemple, un mécanisme ou une implication clairement nouvelle.
\end{quote}

\paragraph{Russian}
\begin{otherlanguage*}{russian}
\begin{quote}\small
Дополнительное правило разнообразия аргументов:\\
В этом ходе введите хотя бы один аргумент, пример, причинный механизм, влияние на заинтересованные стороны или компромисс, которые ранее не появлялись в дебатах. Не ограничивайтесь повторением формулировки вашего партнёра или предыдущего выступающего. Если вы ссылаетесь на ранее высказанный пункт, расширьте его явно новой причиной, новым примером, новым механизмом или новым следствием.
\end{quote}
\end{otherlanguage*}

\paragraph{Arabic}
\begin{arabicblock}
\begin{quote}\small
قاعدة إضافية لتنوع الحجج:\\
في هذا الدور، اطرح على الأقل حجة أو مثالاً أو آلية سببية أو تأثيراً على أصحاب المصلحة أو مفاضلة لم تظهر سابقاً في النقاش. لا تكتفِ بإعادة صياغة طرح زميلك أو طرح المتحدث السابق. وإذا أشرت إلى نقطة سابقة، فوسّعها بسبب جديد أو مثال جديد أو آلية جديدة أو تأثير جديد بوضوح.
\end{quote}

\end{arabicblock}

\subsection{Prompt-Ablation Condition Definitions}
\label{app:prompt-ablation-prompts}
All conditions use the same motion, target-language, role, side, turn label, prior speech, debate history, and sentence-format placeholders as Appendix~\ref{app:baseline-prompt}. Table~\ref{tab:prompt-condition-definitions} gives the condition-specific instruction content.

\begin{table*}[t]
\centering
\small
\begin{tabular}{lp{0.76\textwidth}}
\toprule
Condition & Operational instruction \\
\midrule
Baseline & Full baseline rules in Appendix~\ref{app:baseline-prompt}; no added novelty rule. \\
Diversity-aware & Full baseline plus the language-matched diversity rule in Appendix~\ref{app:diversity-prompt}. \\
P1 paraphrase & Rewords the baseline opening and all eight rules, keeping the same role, response-to-context, reasoning, formatting, fabrication, concision, and length requirements; no new substantive requirement. \\
P2 minimal & Keeps the baseline opening and reduces the eight rules to three (defend the assigned side, respond to prior context, no markdown), leaving the sentence and length format untouched. \\
P3 strong-novelty & Baseline format plus an explicit requirement to advance at least one argument not yet raised anywhere in the debate and not merely reframe an earlier point. \\
P4 targeted-clash & Baseline format plus a requirement to identify the most important unresolved disagreement between the two sides, address that specific clash, and add at least one previously unraised argument. \\
\bottomrule
\end{tabular}
\caption{Condition-specific prompt content in the six-strategy ablation. Verbatim text for every condition is in Appendix~\ref{app:prompt-ablation-prompts}.}
\label{tab:prompt-condition-definitions}
\end{table*}

We give the full text of each condition below. As in the main runs, the prompt scaffold is written in English for every session and the target language is set by the \texttt{\{language\_name\}} placeholder; only the P3 and P4 inserted rules are rendered in the session's language, and because the ablation covers only English and Chinese, those two rules have English and Chinese versions. The generation script rebuilds the baseline template from the same blocks used by the main run and asserts at startup that the reconstruction is byte-identical to it, so each new condition differs from baseline only by the text shown here.

\paragraph{P1 paraphrase.}
Replaces the baseline opening and \texttt{Rules:} block. Everything from \texttt{Length:} onward is unchanged.
\begin{PromptBlock}
You are taking part in a formal debate on the following motion:
"{motion_text}"
Write exclusively in {language_name}.
You have been assigned the role of {role}, arguing for the {side} side.
The current stage of the debate is: {turn_label}.
Guidelines:
- Argue in favour of the position you have been assigned.
- Engage with the previous speaker's points whenever earlier speeches exist.
- Build your case with clear, well-structured reasoning.
- Never state or describe your assigned role.
- Do not open with greetings or address the audience directly.
- Write in continuous prose only, avoiding bullet points, numbered lists, headings, and markdown of any kind.
- Never invent statistics or figures.
- Stay brief and focused.
\end{PromptBlock}

\paragraph{P2 minimal.}
Keeps the baseline opening and replaces the eight-item \texttt{Rules:} block with exactly the following three rules. Everything from \texttt{Length:} onward is unchanged.
\begin{PromptBlock}
Rules:
- Defend your assigned side.
- Respond to the previous speaker when previous context exists.
- Do not use bullet points, numbered lists, headings, or markdown.
\end{PromptBlock}

\paragraph{P3 strong-novelty.}
Inserted after the baseline \texttt{Rules:} block and before \texttt{Length:}, which is the same insertion point used by the diversity-aware rule in the main runs.
\begin{PromptBlock}
Additional novelty rule:
Before writing, silently review the entire debate history and identify every distinct argument, example, causal mechanism, stakeholder impact, and tradeoff that any speaker has already made. Do not restate any of them, even in different wording. Your turn must introduce at least one argument, example, causal mechanism, stakeholder impact, or tradeoff that has not appeared earlier in the debate. If you engage an earlier point, extend it with a clearly new reason, example, mechanism, or implication. Do not list or summarize the earlier arguments in your speech.
\end{PromptBlock}
{\cjkfont
\begin{quote}\small
额外的论点新颖性规则：\\
在写作之前，请先在心中回顾整场辩论的\\
历史，识别任何发言者已经提出过的\\
每一个论点、例子、因果机制、\\
利益相关方影响和权衡。\\
不要重述其中任何一项，\\
即使换用不同的措辞也不行。\\
你的本轮发言必须提出至少一个\\
此前辩论中没有出现过的论点、\\
例子、因果机制、利益相关方影响\\
或权衡。如果你回应此前的某个观点，\\
必须用一个明显新的理由、例子、\\
机制或影响来扩展它。\\
不要在发言中罗列或总结已有论点。
\end{quote}
}

\paragraph{P4 targeted-clash.}
Inserted at the same point as P3.
\begin{PromptBlock}
Additional engagement rule:
Identify the most important point of disagreement between the two sides that remains unresolved in the debate so far. Advance the debate on that specific point of clash by offering at least one reason, example, causal mechanism, stakeholder impact, or tradeoff that no speaker has raised before.
\end{PromptBlock}
{\cjkfont
\begin{quote}\small
额外的交锋规则：\\
请找出到目前为止双方之间最重要\\
且尚未解决的分歧点。\\
围绕这一具体交锋点推进辩论，\\
提出至少一个此前任何发言者\\
都没有提出过的理由、例子、\\
因果机制、利益相关方影响或权衡。
\end{quote}
}

\section{Argument Extraction Details}
\label{app:argument-schema}

This appendix documents the argument-unit extraction procedure used in Section~\ref{sec:argument-extraction}. It includes the full extraction prompt (Appendix~\ref{app:extract-prompt}), the argument-type label definitions (Appendix~\ref{app:schema-defs}), and the provenance and label-collapsing rule used for downstream analyses (Appendix~\ref{app:schema-provenance}).

\subsection{Extraction Prompt}
\label{app:extract-prompt}

The extractor is invoked once per debate turn with the template below. The placeholders \texttt{\{motion\}} and \texttt{\{turn\_text\}} are filled with the motion in the target language and the agent's raw response for that turn; \texttt{\{side\}} is either \texttt{Pro} or \texttt{Con}. The system prompt instructs the model to emit exactly one JSON object and nothing else. Decoding uses temperature~0 and a 1024-token limit.

\begin{PromptBlock}
You are extracting argument units from a debate turn.
Return JSON only.

Motion: {motion}
Speaker side: {side}
Turn text: {turn_text}

Extract 1 to 4 argument units. Each argument unit should contain a distinct claim or reason.

Schema:
{"arguments": [
  {"claim": "...",
   "warrant": "...",
   "evidence_or_example": "...",
   "argument_type": "causal|moral|institutional|practical|stakeholder|historical|economic|legal|safety|other|unclear"}
]}

Rules:
- Preserve the original language where possible.
- Do not invent arguments not present in the text.
- Split repeated or overlapping points only when they are genuinely distinct.
- If the turn contains only one argument, return one argument.
\end{PromptBlock}

\subsection{Argument-Type Labels}
\label{app:schema-defs}

Table~\ref{tab:argument-schema} lists the label set. The schema contains nine substantive labels, plus \textsc{other} and \textsc{unclear}. These labels are intended as broad categories of justification structure, not topical domains. A single argument may plausibly fit more than one label; the extractor is instructed to choose the single best fit.

\begin{table*}[!htbp]
\centering
\small
\begin{tabular}{lp{0.74\linewidth}}
\toprule
Type & Intended content \\
\midrule
causal & A mechanism, chain of consequences, or claim about how one condition produces another. \\
moral & A claim grounded in fairness, rights, duties, harm, legitimacy, or ethical evaluation. \\
institutional & A claim about governments, courts, schools, firms, platforms, procedures, or organizational capacity. \\
practical & A feasibility, implementation, cost, administrative, or operational argument. \\
stakeholder & A claim centered on effects for a group, constituency, user population, or affected party. \\
historical & An analogy, precedent, historical trajectory, or past case used as evidence. \\
economic & A claim about incentives, markets, prices, productivity, fiscal effects, or resource distribution. \\
legal & A claim about law, regulation, legal rights, liability, enforceability, or constitutional constraints. \\
safety & A claim about risk, security, harm prevention, reliability, or protection from adverse outcomes. \\
other & Residual bucket for arguments outside the nine substantive categories, unclear arguments, off-schema labels, and compound outputs. \\
\bottomrule
\end{tabular}
\caption{Argument-type schema.}
\label{tab:argument-schema}
\par\vspace{2pt}
\parbox{\linewidth}{\raggedright\footnotesize \textit{Note.} The schema is used by the extraction prompt.}
\end{table*}

\subsection{Provenance and Label-Collapsing Rule}
\label{app:schema-provenance}

\paragraph{Provenance.}
The schema was hand-designed by the authors to cover recurring types of justification observed in competitive policy debate. Each label corresponds to a distinct mode of warrant rather than to a topical domain. For example, the same motion can attract \textsc{causal}, \textsc{moral}, and \textsc{economic} arguments simultaneously. The schema is not derived from a published argumentation taxonomy; we use it only as a coarse stratification for the diagnostic analyses in Section~\ref{sec:argument-extraction}, not as a contribution of this paper.

\paragraph{Label collapsing.}
Although the extraction prompt restricts \texttt{argument\_type} to the eleven values above, the extractor occasionally produces off-schema labels such as \texttt{logical}, \texttt{comparative}, \texttt{philosophical}, or \texttt{empirical}, as well as compound forms such as \texttt{historical|economic}. For all downstream analyses, we collapse the label space to ten buckets: the nine substantive labels listed in Appendix~\ref{app:schema-defs}, plus a single \textsc{other} bucket. This bucket absorbs the prompt's \textsc{other} and \textsc{unclear} values, together with all off-schema and compound outputs. The same collapsing rule is applied uniformly across all languages, so cross-lingual comparisons of argument-type distributions use the same ten-way partition.

\paragraph{Extraction reliability.}
Argument extraction is performed by Llama-4-Scout-17B, served as a 4-bit quantized model through vLLM 0.16. The extractor is prompted in a strict-JSON format and returns 1--4 argument units per turn. Across 7{,}456 turns, the extractor produced two unparseable responses (0.027\%). These cases are retained in the raw archive and excluded from the flattened argument-unit table used for similarity scoring. The extractor's argument-type labels are used only for taxonomy-balance analyses; the prior-argument-similarity score itself is computed from semantic similarity between extracted argument units and does not depend on the assigned type label.

\section{Robustness Diagnostics}
\label{app:robustness-diagnostics}

\subsection{Bootstrap Procedure}
\label{app:bootstrap}

All reported paired language contrasts use motion as the resampling cluster. For each bootstrap replicate, we sample motions with replacement and recompute the target contrast on the resampled set. Reported 95\% confidence intervals are percentile intervals, defined by the 2.5th and 97.5th quantiles of the bootstrap distribution. A confidence interval is treated as statistically distinguishable from zero when the 95\% percentile interval excludes zero.

\subsection{Regression Robustness}
\label{app:regression-robustness}

To complement the motion-cluster bootstrap, we fit turn-level regression models on the baseline English--Chinese subset. The dependent variable is per-turn prior-argument similarity, and turn~1 is excluded because it has no prior argument pool. The primary regression robustness specification is OLS with motion fixed effects and standard errors clustered by motion, controlling for turn label, model agent, role, and side. We also fit mixed-effects models with a motion random intercept and the same fixed effects.

The regression estimates in Table~\ref{tab:zhen-regression-robustness} keep the Chinese coefficient positive under all three embeddings and numerically identical to the cluster-bootstrap estimate to four decimals. The mixed-effects models estimate singular random-effect covariance, indicating that the motion random-intercept variance is on the boundary after conditioning on the fixed effects. We therefore use the OLS fixed-effect specification as the primary regression robustness result and treat the mixed-effects rows as an agreement diagnostic.

\begin{table*}[!htbp]
\centering
\small
\setlength{\tabcolsep}{5pt}
\begin{tabular}{llccc}
\toprule
Embedding & Estimator & Coef. (ZH$-$EN) & SE & 95\% CI \\
\midrule
BGE-M3 & Bootstrap & $+0.0464$ & -- & $[+0.0399,+0.0525]$ \\
BGE-M3 & OLS-FE & $+0.0464$ & $0.0033$ & $[+0.0399,+0.0530]$ \\
BGE-M3 & LMM (singular RE) & $+0.0464$ & $0.0027$ & $[+0.0412,+0.0516]$ \\
\midrule
mE5-large & Bootstrap & $+0.0529$ & -- & $[+0.0509,+0.0549]$ \\
mE5-large & OLS-FE & $+0.0529$ & $0.0011$ & $[+0.0508,+0.0550]$ \\
mE5-large & LMM (singular RE) & $+0.0529$ & $0.0009$ & $[+0.0511,+0.0547]$ \\
\midrule
LaBSE & Bootstrap & $+0.0469$ & -- & $[+0.0376,+0.0558]$ \\
LaBSE & OLS-FE & $+0.0469$ & $0.0048$ & $[+0.0374,+0.0563]$ \\
LaBSE & LMM (singular RE) & $+0.0469$ & $0.0040$ & $[+0.0390,+0.0548]$ \\
\bottomrule
\end{tabular}
\caption{Regression robustness for the Chinese--English gap.}
\label{tab:zhen-regression-robustness}
\par\vspace{2pt}
\parbox{\linewidth}{\raggedright\footnotesize \textit{Note.} OLS-FE uses motion fixed effects and motion-clustered standard errors, controlling for turn label, model agent, role, and side. LMM uses the same fixed effects with a motion random intercept; all LMM fits estimate singular random-effect covariance, so these rows are reported as agreement diagnostics. Bootstrap rows repeat the motion-cluster bootstrap estimate for comparison.}
\end{table*}

\subsection{Decoding Diagnostics}
\label{app:decoding-diagnostics}

The main experiment uses provider-default decoding because not all model providers expose comparable decoding controls. To test whether the observed language gap could be explained by sampling settings, we conduct fixed-context diagnostics on the two model agents that expose decoding controls: DeepSeek-V3.2 and Mistral Large 3.

For temperature, we generate single-turn continuations at $T \in \{0.2, 0.7, 1.0\}$. For nucleus sampling, we generate continuations at $p \in \{0.70, 0.85, 0.95, 1.00\}$. These diagnostics are not treated as replacements for the full debate experiment. They are used to assess whether plausible decoding choices systematically remove or reverse the Chinese--English prior-argument-similarity gap.

Across all model--language--embedding cells in these diagnostics, the within-model decoding effect is $|\Delta| \leq 0.02$ absolute. This is smaller than the cross-lingual gap of roughly $+0.04$ to $+0.08$ observed in the corresponding language contrasts. Most decoding-contrast bootstrap confidence intervals overlap zero, while the language-contrast confidence intervals exclude zero. We therefore interpret the decoding diagnostics as evidence that the tested sampling settings do not systematically account for the observed Chinese--English prior-argument-similarity gap.

\subsection{Translate-and-Rescore Decomposition}
\label{app:translate-rescore}

This probe asks how much of the Chinese--English gap travels with the surface language. We use the same stratified 30-motion English--Chinese subset as the second-extractor check. Chinese baseline turns are machine-translated into English with Google Translate (accessed 2026-07-09), then re-extracted with the locked Llama-4-Scout prompt and schema and rescored entirely in English embedding space; the extractor receives the English original motion. To control for translation-induced paraphrase noise, native English sessions undergo an English$\rightarrow$Chinese$\rightarrow$English round trip before the same re-extraction and rescoring procedure. Turn~1 is excluded, and inference uses 5{,}000 motion-cluster bootstrap resamples with seed 20260709.

\begin{table*}[t]
\centering
\small
\begin{tabular}{lcc}
\toprule
Comparison & Gap & 95\% CI \\
\midrule
Native Chinese $-$ native English, BGE-M3 & $+0.0472$ & $[+0.0382,+0.0563]$ \\
MT(Chinese$\rightarrow$English) $-$ native English, BGE-M3 & $+0.0085$ & $[+0.0017,+0.0152]$ \\
MT(Chinese$\rightarrow$English) $-$ round-trip English, BGE-M3 & $+0.0140$ & $[+0.0057,+0.0223]$ \\
Round-trip English $-$ native English, BGE-M3 & $-0.0055$ & $[-0.0109,-0.0002]$ \\
\midrule
MT Chinese $-$ round-trip English, multilingual-E5-large & $+0.0033$ & $[+0.0001,+0.0065]$ \\
MT Chinese $-$ round-trip English, LaBSE & $+0.0151$ & $[+0.0040,+0.0257]$ \\
\bottomrule
\end{tabular}
\caption{Translate-and-rescore decomposition on the 30-motion subset.}
\label{tab:translate-rescore}
\end{table*}

Translation into English removes most of the native BGE-M3 gap, which implicates surface language or representation geometry. The treatment-minus-round-trip contrast nevertheless remains positive under all three embeddings, which is consistent with a smaller content-level component. This analysis does not identify whether that component originates in pretraining, post-training, tokenization during generation, discourse conventions, or their interaction.

\subsection{Sentence-Count Compliance Audit}
\label{app:sentence-compliance}

We audit sentence-count compliance for all 6{,}816 turns in the baseline and diversity-prompt experiments. Opening, rebuttal, and cross-examination turns are expected to contain five sentences, while closing turns are expected to contain four sentences. Sentence segmentation uses language-specific sentence terminators; sentence length is measured in whitespace tokens for English, Arabic, Spanish, French, and Russian, and in non-whitespace characters for Chinese.

The compliance audit in Table~\ref{tab:sentence-compliance-summary} indicates that baseline compliance is high across languages and does not single out Chinese. English and Chinese have similar exact-compliance rates (95.6\% vs. 94.4\%) and nearly identical mean sentence counts (4.84 vs. 4.83). The largest baseline over-count rates occur in Spanish and French rather than Chinese. In the diversity-prompt condition, compliance decreases modestly for several languages and sentence lengths increase, consistent with more elaborated turns under the intervention.

We also test whether sentence-count deviations are associated with the similarity metric or with extraction volume. In the baseline condition, the absolute sentence-count deviation has weak Spearman correlations with BGE-M3 prior-argument similarity in English ($\rho=0.129$) and Chinese ($\rho=0.087$), and with extracted-argument counts in English ($\rho=0.049$) and Chinese ($\rho=0.071$). Observed sentence count is negatively correlated with prior-argument similarity in both languages (English $\rho=-0.315$, Chinese $\rho=-0.270$), so this pattern is not a Chinese-specific advantage. Because Chinese length is measured in characters while the other languages use whitespace tokens, we do not interpret pooled cross-language length correlations.

\begin{table*}[!htbp]
\centering
\small
\setlength{\tabcolsep}{4pt}
\begin{tabular}{llrrrrrrrr}
\toprule
Cond. & Lang. & Turns & Exact & Under & Over & Mean sents. & Mean len. & Mean args & Mean sim. \\
\midrule
Base & ar & 568 & 0.979 & 0.002 & 0.019 & 4.78 & 19.98 & 3.86 & 0.637 \\
Base & en & 568 & 0.956 & 0.000 & 0.044 & 4.84 & 21.07 & 3.83 & 0.662 \\
Base & es & 568 & 0.884 & 0.002 & 0.114 & 5.18 & 22.78 & 3.83 & 0.641 \\
Base & fr & 568 & 0.917 & 0.004 & 0.079 & 4.99 & 23.36 & 3.82 & 0.644 \\
Base & ru & 568 & 0.963 & 0.000 & 0.037 & 4.81 & 19.91 & 3.84 & 0.652 \\
Base & zh & 568 & 0.944 & 0.000 & 0.056 & 4.83 & 42.08 & 3.92 & 0.708 \\
\midrule
Div. & ar & 568 & 0.949 & 0.016 & 0.035 & 4.82 & 25.14 & 3.90 & 0.621 \\
Div. & en & 568 & 0.933 & 0.000 & 0.067 & 4.89 & 25.39 & 3.85 & 0.638 \\
Div. & es & 568 & 0.894 & 0.000 & 0.106 & 5.17 & 27.75 & 3.92 & 0.620 \\
Div. & fr & 568 & 0.887 & 0.005 & 0.107 & 5.16 & 29.28 & 3.85 & 0.619 \\
Div. & ru & 568 & 0.968 & 0.002 & 0.030 & 4.86 & 24.86 & 3.86 & 0.630 \\
Div. & zh & 568 & 0.900 & 0.000 & 0.100 & 4.89 & 54.50 & 3.91 & 0.688 \\
\bottomrule
\end{tabular}
\caption{Sentence-count compliance by language and prompt condition.}
\label{tab:sentence-compliance-summary}
\par\vspace{2pt}
\parbox{\linewidth}{\raggedright\footnotesize \textit{Note.} Exact compliance is measured against the prompt template: five sentences for opening, rebuttal, and cross-examination turns, and four sentences for closing turns. Sentence length is in whitespace tokens for English, Arabic, Spanish, French, and Russian, and in non-whitespace characters for Chinese. Mean sim. is BGE-M3 prior-argument similarity.}
\end{table*}

\subsection{Diversity-Prompt Intervention Design}
\label{app:diversity-intervention-design}

The diversity-aware prompt appends one paragraph to the baseline debate prompt. It instructs agents to avoid merely restating earlier arguments and to add new considerations, examples, causal mechanisms, stakeholder impacts, or tradeoffs where possible. The intervention was originally planned for English and Chinese, then expanded to all six languages. The English and Chinese diversity prompts follow the planned wording; the Arabic, Spanish, French, and Russian versions were authored to mirror the same structure and targets.

This intervention is interpreted narrowly. It tests whether this particular surface-level diversity instruction reduces prior-argument similarity and whether it narrows the Chinese--English gap. It does not test all possible anti-repetition prompts, debate formats, or instruction-tuning interventions.

\subsection{Prompt-Ablation Design and Diagnostics}
\label{app:prompt-ablation-design}
The broader prompt ablation compares the six conditions in Table~\ref{tab:prompt-condition-definitions} on the same 10 motions used for repeat stability. Each motion--language--condition cell has three independent generations. Baseline and diversity-aware generations reuse the original and two repeat runs; P1--P4 are newly generated. The same role-to-model mapping is held fixed within motion across conditions and languages. Extraction uses the original Llama-4-Scout pipeline at temperature~0, and the outcome remains session-level mean maximum similarity to earlier arguments.

For absolute-level effects, each condition is compared with baseline within language and motion. For gap interactions, we first compute the Chinese--English gap within each condition and motion, then difference that gap from baseline before bootstrapping motions. This preserves the paired design and avoids treating turns or arguments as independent. Format diagnostics include extracted arguments per turn and exact sentence-count compliance.

\subsection{Repeat-Stability Subset}
\label{app:repeat-stability}

To evaluate generation-level stability, we select a 10-motion subset and generate two additional repeats for English and Chinese under both baseline and diversity-prompt conditions. Combined with the original run, this yields three independent generations per cell. The subset includes motions that initially showed strong, medium, and weak or counterexample Chinese--English gaps, allowing us to distinguish stable aggregate effects from unstable motion-level rankings.

We use this analysis only for stability assessment. Main claims remain based on the full 71-motion baseline and diversity experiments. The aggregate language-level patterns are stable across repeats, but individual motion rankings fluctuate. In particular, M053 and M061 appear weak or counterexample-like in the first run but become positive when averaged over three repeats. In the opposite direction, M013 shows the strongest first-run motion-level Chinese--English effect under BGE-M3 ($+0.102$) but attenuates to $+0.020$ and $+0.024$ in the second and third runs. These movements support treating per-motion rankings as exploratory appendix material rather than as part of the main inferential claims.

\subsection{Embedding Representations and Repeat Noise}
\label{app:embedding-noise}

The main analysis reports prior-argument-similarity estimates separately under BGE-M3, multilingual-E5-large, and LaBSE rather than collapsing them into a single representation space. BGE-M3 was selected before the main experiment because of its broad multilingual coverage; multilingual-E5-large and LaBSE are used as independent robustness checks.

The repeat-stability analysis also provides a scale for interpreting effect sizes. The within-cell standard deviation over three independent generations is 0.009 for BGE-M3, 0.028 for multilingual-E5-large, and 0.043 for LaBSE. Thus, the pooled Chinese--English language gap is larger than the repeat-generation noise for BGE-M3 and multilingual-E5-large, while LaBSE has the highest within-cell noise among the three embeddings. This is relevant when interpreting the diversity-prompt gap-reduction estimate under LaBSE: its directional reduction is comparable in magnitude to the repeat noise of that embedding and its confidence interval includes zero.

\section{Metric Diagnostics and Manual Calibration}
\label{app:manual-audit-protocol}

This appendix reports the additional metric diagnostics and manual calibration analyses summarized in Section~\ref{sec:res-robustness} and Section~\ref{sec:res-manual-calibration}. These analyses qualify the interpretation of prior-argument similarity as an aggregate diagnostic rather than as a direct item-level human label.

\subsection{Metric Diagnostic Variants}
\label{app:metric-diagnostics}

The main-text diagnostics in Table~\ref{tab:metric-diagnostics} include the Chinese--English gap under null adjustment, $z$-scoring, fixed-size prior pools, and input-field ablations. The fixed-$K$ and input-field variants preserve the positive direction across all three embedding models. Null adjustment is more sensitive to embedding geometry: BGE-M3 and LaBSE remain positive, while multilingual-E5-large flips sign after subtracting its language-specific null baseline. We therefore use these diagnostics to characterize metric sensitivity rather than to claim that every calibrated variant gives the same result.

\subsection{Within-Language Embedding Agreement and Language-Specific Calibration}
\label{app:embedding-calibration}
Raw language means differ across embedding geometries, but the embeddings agree strongly on which turns are relatively high or low within each language. Table~\ref{tab:embedding-rank-agreement} reports the range of the three pairwise turn-level Spearman correlations for each language.

\begin{table}[t]
\centering
\small
\begin{tabular}{lc}
\toprule
Language & Pairwise Spearman range \\
\midrule
English & $0.766$--$0.789$ \\
Arabic  & $0.769$--$0.830$ \\
Spanish & $0.741$--$0.797$ \\
French  & $0.702$--$0.766$ \\
Russian & $0.749$--$0.802$ \\
Chinese & $0.696$--$0.812$ \\
\bottomrule
\end{tabular}
\caption{Turn-level within-language Spearman range across BGE-M3, multilingual-E5-large, and LaBSE.}
\label{tab:embedding-rank-agreement}
\end{table}

The lowest single correlation occurs in Chinese, so cross-embedding agreement is not uniquely high for the focal language. Figure~\ref{fig:crosslingual-landscape} mainly reflects embedding-specific absolute offsets across languages. To test sensitivity to those offsets, we standardize each observed argument-level score against a same-language, different-motion, same-turn-position null distribution of maximum similarities. Table~\ref{tab:zcal-language-gaps} shows that Chinese is the only language with a positive and significant $z$-scored gap from English under all three embeddings.

\begin{table*}[t]
\centering
\small
\begin{tabular}{lrrr}
\toprule
Language $-$ English & BGE-M3 & multilingual-E5-large & LaBSE \\
\midrule
Arabic  & $-0.59^{*}$ & $-0.34^{*}$ & $-0.18$ \\
Spanish & $-0.14$ & $-0.06$ & $-0.06$ \\
French  & $-0.33^{*}$ & $-0.56^{*}$ & $+0.06$ \\
Russian & $-0.11$ & $-0.02$ & $+0.23^{*}$ \\
Chinese & $+0.52^{*}$ & $+1.50^{*}$ & $+0.39^{*}$ \\
\bottomrule
\end{tabular}
\caption{Language-minus-English gaps after language-specific $z$ calibration.}
\label{tab:zcal-language-gaps}
\par\vspace{2pt}
\parbox{\linewidth}{\raggedright\footnotesize \textit{Note.} Asterisks indicate a 95\% motion-cluster bootstrap CI excluding zero. Null calibration can absorb corpus-level signal as well as representation offsets, so these values are sensitivity checks rather than replacements for the raw primary estimates. In particular, the paper makes no substantive raw-score English-versus-Arabic, Spanish, French, or Russian claim.}
\end{table*}

\subsection{Cross-Encoder Tail Rescoring}
\label{app:cross-encoder-tail}

To test whether the high-similarity tail is specific to BGE-M3 bi-encoder cosine geometry, we rescore all English--Chinese BGE-M3 max-prior argument pairs with the multilingual cross-encoder reranker \texttt{BAAI/bge-reranker-v2-m3}. This check does not rerun all pairwise nearest-neighbor search; it asks whether the same current--nearest-prior pairs remain disproportionately Chinese in a high-score tail under an orthogonal pair scorer. Table~\ref{tab:cross-encoder-tail} shows that Chinese remains overrepresented in the top 20\%, top 10\%, and top 5\% tails, with smaller effects than under BGE-M3.

\begin{table*}[!htbp]
\centering
\small
\setlength{\tabcolsep}{5pt}
\begin{tabular}{llrrrr}
\toprule
Scorer & Tail & EN rate & ZH rate & ZH--EN & 95\% motion-cluster CI \\
\midrule
BGE-M3 bi-encoder & top 20\% & 0.111 & 0.287 & $+0.176$ & $[+0.150,+0.203]$ \\
BGE reranker & top 20\% & 0.154 & 0.246 & $+0.092$ & $[+0.067,+0.119]$ \\
BGE-M3 bi-encoder & top 10\% & 0.063 & 0.137 & $+0.074$ & $[+0.052,+0.096]$ \\
BGE reranker & top 10\% & 0.076 & 0.124 & $+0.048$ & $[+0.030,+0.066]$ \\
BGE-M3 bi-encoder & top 5\% & 0.029 & 0.070 & $+0.041$ & $[+0.026,+0.056]$ \\
BGE reranker & top 5\% & 0.040 & 0.059 & $+0.019$ & $[+0.007,+0.033]$ \\
\bottomrule
\end{tabular}
\caption{High-overlap tail rates under BGE-M3 and cross-encoder reranker scores.}
\label{tab:cross-encoder-tail}
\par\vspace{2pt}
\parbox{\textwidth}{\raggedright\footnotesize \textit{Note.} We rescore the 3{,}856 English--Chinese baseline BGE-M3 max-prior pairs with \texttt{BAAI/bge-reranker-v2-m3}. Thresholds are pooled English--Chinese percentiles of each scorer. The reranker is an orthogonal scoring check, not a human repetition label.}
\end{table*}

\subsection{Automatic Extraction Diagnostics}
\label{app:extraction-diagnostics}
Automatic extraction diagnostics by language, including extracted arguments per turn, claim length, empty warrant and evidence rates, and off-schema type rates, are summarized in Table~\ref{tab:extraction-diagnostics}. These checks characterize extraction artifacts that could affect prior-argument similarity and motivate the manual extraction audit below.

\begin{table*}[!htbp]
\centering
\small
\begin{tabular}{lcccccc}
\toprule
Diagnostic & en & ar & es & fr & ru & zh \\
\midrule
Arguments/turn & 3.83 & 3.86 & 3.83 & 3.82 & 3.84 & 3.92 \\
Claim characters & 88 & 66 & 96 & 96 & 86 & 27 \\
Empty warrant & 31.7\% & 28.9\% & 25.5\% & 27.8\% & 29.4\% & 45.0\% \\
Empty evidence & 69.3\% & 76.3\% & 69.2\% & 72.8\% & 70.9\% & 77.9\% \\
Off-schema type & 2.3\% & 2.1\% & 2.1\% & 2.4\% & 2.2\% & 3.9\% \\
\bottomrule
\end{tabular}
\caption{Automatic extraction diagnostics by language.}
\label{tab:extraction-diagnostics}
\end{table*}

\subsection{Extraction-Length and Field-Completeness Controls}
\label{app:extraction-controls}

Chinese extracted claims are shorter on average than English claims, and the original extractor leaves the Chinese \texttt{warrant} field empty more often. To test whether these extraction features mechanically account for the Chinese--English gap, we fit turn-level OLS models on baseline English--Chinese turns with non-empty prior pools. The full specification controls for mean claim length, argument count, empty warrant and evidence rates, prior-pool size, sentence count, turn label, model agent, role, side, and motion fixed effects, with standard errors clustered by motion. Table~\ref{tab:zhen-extraction-controls} shows that the adjusted Chinese coefficient remains positive under all three embedding models.

\begin{table*}[!htbp]
\centering
\small
\setlength{\tabcolsep}{3.5pt}
\begin{tabularx}{\textwidth}{ll>{\raggedright\arraybackslash}Xrrrr}
\toprule
Embedding & Spec. & Controls & Coef. & SE & 95\% CI & $N$ \\
\midrule
BGE-M3 & A & none & $+0.0464$ & 0.0032 & $[+0.0401,+0.0527]$ & 994 \\
BGE-M3 & B & turn, agent, role, side, motion FE & $+0.0464$ & 0.0033 & $[+0.0399,+0.0530]$ & 994 \\
BGE-M3 & C & B + claim length, argument count, empty warrant/evidence rates, prior-pool size, sentence count & $+0.0431$ & 0.0064 & $[+0.0304,+0.0557]$ & 994 \\
BGE-M3 & C boot. & same as C, motion-cluster bootstrap & $+0.0430$ & 0.0061 & $[+0.0313,+0.0551]$ & 994 \\
mE5-large & C & full controls as in C & $+0.0470$ & 0.0026 & $[+0.0418,+0.0521]$ & 994 \\
LaBSE & C & full controls as in C & $+0.0240$ & 0.0098 & $[+0.0047,+0.0433]$ & 994 \\
\bottomrule
\end{tabularx}
\caption{Chinese--English prior-argument-similarity gap under extraction-length and field-completeness controls.}
\label{tab:zhen-extraction-controls}
\par\vspace{2pt}
\parbox{\textwidth}{\raggedright\footnotesize \textit{Note.} The outcome is turn-level prior-argument similarity on baseline English--Chinese turns with non-empty prior pools. Coefficients are Chinese minus English. Specifications A--C use OLS with motion-clustered standard errors; C boot. resamples motions with replacement. All rows use 71 motions.}
\end{table*}

\subsection{Second-Extractor Robustness}
\label{app:second-extractor}

To test dependence on the single original extractor, we re-extract a 30-motion English--Chinese subset with DeepSeek-Chat at temperature~0, using the same argument-unit schema. The subset is stratified by the original BGE-M3 motion-level Chinese--English gap, with 10 motions from each tertile. This is a targeted extractor-dependence check rather than a full re-run. Table~\ref{tab:second-extractor-robustness} shows that the BGE-M3 Chinese--English gap remains positive and similar in magnitude under the second extractor, even though the second extractor fills warrant and evidence fields much more aggressively.

\begin{table*}[!htbp]
\centering
\small
\setlength{\tabcolsep}{3pt}
\begin{tabular}{lrrrrrrr}
\toprule
Extractor & Motions & Args/turn EN & Args/turn ZH & Empty W EN & Empty W ZH & Claim chars EN & Claim chars ZH \\
\midrule
Llama-4-Scout & 30 & 3.84 & 3.92 & 0.326 & 0.464 & 89.5 & 27.1 \\
DeepSeek-Chat & 30 & 3.59 & 3.82 & 0.013 & 0.011 & 120.2 & 35.1 \\
\bottomrule
\end{tabular}
\par\vspace{3pt}
\begin{tabular}{lcc}
\toprule
Extractor & BGE-M3 ZH--EN gap & 95\% motion-cluster CI \\
\midrule
Llama-4-Scout & $+0.0472$ & $[+0.0379,+0.0568]$ \\
DeepSeek-Chat & $+0.0506$ & $[+0.0419,+0.0593]$ \\
\bottomrule
\end{tabular}
\caption{Second-extractor robustness on a 30-motion English--Chinese subset.}
\label{tab:second-extractor-robustness}
\par\vspace{2pt}
\parbox{\textwidth}{\raggedright\footnotesize \textit{Note.} The subset includes 10 motions from each tertile of the original BGE-M3 Chinese--English motion-level gap. Each extractor uses the same JSON schema and all eight turns for English and Chinese. Empty W is the fraction of extracted arguments with an empty \texttt{warrant} field.}
\end{table*}

\subsection{Manual Extraction Audit}
\label{app:manual-extraction-audit}

Annotators audited English and Chinese turns for extraction quality, including whether the extracted units captured all major arguments, whether the boundaries were over- or under-split, whether hallucinated arguments were present, and the overall extraction quality on a 1--5 scale. Exact agreement was high for most labels; some kappa values are lower because the base rates are extreme, with most extractions judged good or usable.

\begin{table*}[!htbp]
\centering
\small
\begin{tabular}{lccc}
\toprule
Measure & English & Chinese & Difference (95\% CI) \\
\midrule
Overall quality (1--5) & 3.98 & 3.92 & $-0.06$ $[-0.17,+0.04]$ \\
All major arguments captured & 0.988 & 0.959 & $-0.03$ $[-0.09,+0.02]$ \\
Over-split rate & 0.000 & 0.020 & $+0.02$ $[0.00,+0.07]$ \\
Under-split rate & 0.000 & 0.000 & $+0.00$ \\
Hallucination & 0.020 & 0.000 & $-0.02$ $[-0.07,0.00]$ \\
Missing major argument & 0.000 & 0.056 & $+0.056$ $[0.00,+0.143]$ \\
\bottomrule
\end{tabular}
\caption{Manual extraction audit for English and Chinese.}
\label{tab:manual-extraction-audit}
\par\vspace{2pt}
\parbox{\linewidth}{\raggedright\footnotesize \textit{Note.} Differences are Chinese minus English (ZH--EN).}
\end{table*}

\subsection{Manual Repetition Calibration}
\label{app:manual-repetition-calibration}
For perceived repetition, annotators labeled current--prior argument pairs as new, topically related, partial repetition or extension, or substantive repetition. The main binary collapse treats partial repetition and substantive repetition as repeated; the strict collapse uses only substantive repetition. Agreement is high (exact agreement $0.83$, binary $\kappa=0.733$, weighted $\kappa=0.857$), but the automatic score is only weakly aligned with item-level human labels.

To separate a weakness of the bi-encoder from a weakness of automatic scoring in general, we score the same 200 adjudicated pairs with two further instruments: the multilingual cross-encoder reranker and a blinded \texttt{deepseek-chat} pair judge. Table~\ref{tab:instrument-ladder} compares all three against the human labels.

\begin{table*}[t]
\centering
\small
\begin{tabular}{lrrrr}
\toprule
Instrument & $N$ & Spearman $\rho$ & AUC repeated ($\geq2$) & AUC strict ($=3$) \\
\midrule
BGE-M3 cosine & 200 & $0.195$ & $0.591$ & $0.638$ \\
BGE cross-encoder & 200 & $0.255$ & $0.634$ & $0.645$ \\
DeepSeek-Chat pair judge & 200 & $0.553$ & $0.640$ & $0.813$ \\
\bottomrule
\end{tabular}
\caption{Item-level alignment with the same 200 adjudicated argument pairs.}
\label{tab:instrument-ladder}
\par\vspace{2pt}
\parbox{\linewidth}{\raggedright\footnotesize \textit{Note.} All three instruments score the same adjudicated pairs. Higher $\rho$ and strict-repetition AUC support a ranking interpretation, not label interchangeability.}
\end{table*}

The pair judge uses the original codebook, temperature~0, and only the two claims; it is blinded to embedding scores and model identity. Its higher $\rho$ and strict-repetition AUC support a ranking interpretation, not label interchangeability. It assigns ``new'' to 167/200 pairs where the adjudicated human set contains no label-0 items, producing exact four-class agreement $0.045$ and binary $\kappa=0.251$. Its repeated rate is $0.16$ versus $0.55$ for humans, and its strict rate is $0.015$ versus $0.195$. The LLM judge is therefore a substantially more conservative complementary instrument.

The direct labeled subset is stratified and does not show a clear Chinese--English human-label gap. The global-tail analysis instead asks where English and Chinese items fall in the full BGE-M3 similarity distribution and then combines those population rates with the pooled human calibration curve. This analysis shows that Chinese items are substantially overrepresented in the high-similarity tail, and that the tail is enriched for strict substantive repetition.

\begin{table*}[!htbp]
\centering
\small
\begin{tabular}{lccc}
\toprule
Collapse & English weighted & Chinese weighted & ZH--EN (95\% CI) \\
\midrule
Repeated binary ($\{2,3\}$) & 0.557 $[0.463,0.647]$ & 0.551 $[0.456,0.646]$ & $-0.006$ $[-0.139,+0.124]$ \\
Strict repeated ($3$) & 0.211 $[0.133,0.292]$ & 0.174 $[0.113,0.243]$ & $-0.036$ $[-0.138,+0.068]$ \\
\bottomrule
\end{tabular}
\caption{Population-weighted manual repetition rates.}
\label{tab:manual-pop-weighted}
\par\vspace{2pt}
\parbox{\linewidth}{\raggedright\footnotesize \textit{Note.} Rates are from the stratified EN/ZH audit. Direct labeled rates do not show a clear ZH--EN gap.}
\end{table*}

\begin{table*}[!htbp]
\centering
\small
\begin{tabular}{lccc}
\toprule
Analysis & English & Chinese & ZH--EN (95\% CI) \\
\midrule
Top quintile rate & 0.112 $[0.093,0.130]$ & 0.288 $[0.262,0.315]$ & $+0.176$ $[+0.148,+0.204]$ \\
Top decile rate & 0.063 $[0.050,0.078]$ & 0.137 $[0.117,0.157]$ & $+0.074$ $[+0.051,+0.096]$ \\
Calibrated repeated binary & 0.544 $[0.473,0.616]$ & 0.580 $[0.510,0.647]$ & $+0.036$ $[-0.004,+0.075]$ \\
Calibrated strict repeated & 0.184 $[0.132,0.237]$ & 0.237 $[0.171,0.298]$ & $+0.053$ $[+0.015,+0.089]$ \\
\bottomrule
\end{tabular}
\caption{Global-tail calibration for BGE-M3 prior-argument similarity.}
\label{tab:manual-global-tail}
\par\vspace{2pt}
\parbox{\linewidth}{\raggedright\footnotesize \textit{Note.} Global thresholds are computed on the pooled English--Chinese baseline population.}
\end{table*}

\paragraph{Qualitative calibration example.}
One audited pair from a subscription-model debate illustrates the type of high-overlap content captured by the metric. The example comes from motion M015 in the manual repetition audit. The prior claim is from turn~1 by Claude Opus 4.7 as Pro First Speaker, and the current claim is from turn~7 by DeepSeek-V3.2 as Pro Second Speaker.
\begin{quote}\small
\emph{Prior:} {\cjkfont 订阅制削弱了消费者的\\
财产权和选择权}\\
\emph{Later:} {\cjkfont 订阅制的本质是从所有权\\
转向持续租赁，这一转变使\\
消费者对已付费内容的\\
控制权彻底丧失}
\end{quote}
The pair has BGE-M3 similarity $0.778$ and was labeled substantive repetition by both annotators: the surface wording differs, but both claims make the same ownership-loss argument about subscription models.

\paragraph{Additional non-Chinese high-similarity examples.}
High prior-argument similarity is not a Chinese-specific phenomenon; it is a property of the metric that appears in every language. Table~\ref{tab:multilingual-high-sim-examples} gives one pair with similarity above $0.90$ for each of the five non-Chinese languages, glossed into English for compact comparison.

\begin{table*}[t]
\centering
\small
\begin{tabularx}{\textwidth}{lclXX}
\toprule
Language & Motion & Similarity & Prior claim & Current claim \\
\midrule
English & M037 & $0.997$ & art can be assessed objectively & art can be evaluated objectively \\
Spanish & M019 & $0.911$ & specialization is effective without integration & specialization does not need forced integration \\
Russian & M009 & $0.901$ & rapid growth creates jobs, a tax base, and resources for social programs & rapid growth creates jobs \\
Arabic  & M035 & $0.953$ & international trust and legal expertise must not be a reason to exclude the poor & legal expertise and international trust are not pretexts for excluding the poor \\
French  & M003 & $0.952$ & American involvement benefited Panama & U.S.\ involvement in Panama benefited Panama \\
\bottomrule
\end{tabularx}
\caption{Representative high-similarity pairs in all five non-Chinese languages, with English glosses. Verbatim source-language strings are in Appendix~\ref{app:multilingual-verbatim}.}
\label{tab:multilingual-high-sim-examples}
\par\vspace{2pt}
\parbox{\linewidth}{\raggedright\footnotesize \textit{Note.} Similarity is BGE-M3 cosine between each argument's \texttt{claim} and its nearest earlier \texttt{claim} in the same session. English glosses are shown here for compact comparison. These pairs illustrate substantive semantic return rather than language-specific token overlap. Arabic and French additionally contain string-identical pairs at similarity $1.000$ (Arabic M061, French M062 and M069), so exact restatement is not confined to any one language.}
\end{table*}

\subsection{Verbatim Multilingual High-Similarity Pairs}
\label{app:multilingual-verbatim}

For each language we reproduce the source-language strings of the pair glossed in Table~\ref{tab:multilingual-high-sim-examples}, together with the turn positions and speaking agents. Similarity is BGE-M3 cosine between the two \texttt{claim} fields.

\paragraph{English (M037, turn~4 GPT-5.5 $\rightarrow$ turn~6 DeepSeek-V3.2, $0.997$).}
\begin{quote}\small
\emph{Prior:} The quality of art can be assessed objectively\\
\emph{Later:} The quality of art can be evaluated objectively
\end{quote}

\paragraph{Spanish (M019, turn~6 Claude Opus 4.7 $\rightarrow$ turn~8 Mistral Large 3, $0.911$).}
\begin{quote}\small
\emph{Prior:} La especializaci\'on es efectiva sin necesidad de integraci\'on\\
\emph{Later:} La especializaci\'on no necesita integraci\'on forzada para funcionar
\end{quote}

\paragraph{French (M003, turn~3 $\rightarrow$ turn~7, both GPT-5.5, $0.952$).}
\begin{quote}\small
\emph{Prior:} l'implication am\'ericaine a b\'en\'efici\'e au Panama\\
\emph{Later:} l'implication des \'Etats-Unis au Panama a b\'en\'efici\'e au Panama
\end{quote}

\paragraph{Russian (M009, turn~2 GPT-5.5 $\rightarrow$ turn~4 Mistral Large 3, $0.901$).}
\begin{otherlanguage*}{russian}
\begin{quote}\small
\emph{Prior:} Быстрый рост создаёт рабочие места, налоговую базу\\
и ресурсы для социальных программ\\
\emph{Later:} Быстрый рост создаёт рабочие места
\end{quote}
\end{otherlanguage*}

\paragraph{Arabic (M035, turn~7 Claude Opus 4.7 $\rightarrow$ turn~8 DeepSeek-V3.2, $0.953$).}
\begin{arabicblock}
\begin{quote}\small
\emph{Prior:} الثقة الدولية و الخبرة القانونية لا يجب أن تكونا سبباً في إقصاء الفقراء\\
\emph{Later:} الخبرة القانونية والثقة الدولية ليست ذرائع لإقصاء الفقراء
\end{quote}
\end{arabicblock}
\noindent In the Arabic pair the prior claim is made by a Pro speaker and the later claim by a Con speaker, so the two sides converge on nearly the same assertion in different wording. This illustrates why we treat the score as a diagnostic of semantic return rather than of within-side redundancy.

\paragraph{High similarity without redundant restatement.}
The converse also holds: a high-similarity pair need not be redundant. Table~\ref{tab:high-sim-nonredundant} shows audited pairs that both annotators scored as highly similar and that nevertheless function as rebuttal or extension rather than restatement.

\begin{table*}[t]
\centering
\small
\begin{tabularx}{\textwidth}{lcclX}
\toprule
Example & Similarity & Human labels & Function & Explanation \\
\midrule
ZH M006 & $0.759$ & 2/2 & Rebuttal & Reuses the frame of specialized funds and limited-partner risk but reverses its valence. \\
EN M071 & $0.705$ & 2/2 & Rebuttal & Reuses the prior flexibility frame in order to challenge it. \\
ZH M026 & $0.695$ & 2/2 & Extension & Retains the unified-market theme while adding a distinct mechanism. \\
\bottomrule
\end{tabularx}
\caption{Human-audited examples showing why cosine similarity is not an item-level redundancy label.}
\label{tab:high-sim-nonredundant}
\end{table*}

The functional composition of the BGE-M3 top decile makes the same point at scale: 35.0\% of sampled pairs are rebuttal/response, 36.5\% are refinement/extension, and 28.5\% are redundant restatement. Semantic return is therefore a process signal whose local argumentative function requires a second instrument or human reading.

\subsection{Functional Decomposition of High-Similarity Pairs}
\label{app:functional-decomposition}
Finally, we conduct an exploratory functional decomposition of 200 English--Chinese current--nearest-prior pairs sampled from the pooled BGE-M3 top decile. DeepSeek-Chat assigns each pair one primary function label using a fixed prompt and category set; 20 items were manually spot-checked. Table~\ref{tab:function-label-distribution} shows that the top-decile tail is functionally mixed: redundant restatement appears in a non-trivial share of pairs, but most sampled pairs are labeled as rebuttal/response or refinement/extension. We therefore do not interpret high prior-argument similarity as an item-level quality or redundancy label.

\begin{table}[!htbp]
\centering
\small
\begin{tabular}{lrrr}
\toprule
Function label & EN & ZH & Total \\
\midrule
Redundant restatement & 40.0\% & 17.0\% & 28.5\% \\
Rebuttal or response & 28.0\% & 42.0\% & 35.0\% \\
Refinement or extension & 32.0\% & 41.0\% & 36.5\% \\
Quotation or reference & 0.0\% & 0.0\% & 0.0\% \\
Unclear or other & 0.0\% & 0.0\% & 0.0\% \\
\bottomrule
\end{tabular}
\caption{Functional labels for sampled BGE-M3 top-decile pairs.}
\label{tab:function-label-distribution}
\par\vspace{2pt}
\parbox{\linewidth}{\raggedright\footnotesize \textit{Note.} Labels are assigned to 200 English--Chinese current/prior pairs sampled from the pooled BGE-M3 top decile, balanced by language. Labels use DeepSeek-Chat at temperature 0 with a fixed category set and a 20-item manual spot check.}
\end{table}

\section{Process Similarity and Judged Debate Quality}
\label{app:outcome-quality}

This appendix reports the outcome-quality evaluation introduced in Section~\ref{sec:outcome-quality-method} and summarized in Section~\ref{sec:res-outcome-quality}. It asks whether prior-argument similarity carries information about the debate a reader receives, and whether reducing it improves that debate equally across languages.

\subsection{Judge Protocol and Reliability}
Gemini-3.5-Flash at temperature~0 independently scores all 284 English and Chinese sessions from the baseline and diversity-aware conditions. The judge is not one of the four debate agents. Each call contains only the motion and eight role-labeled turns. Build-time checks exclude agent identities, prompt-condition labels, prior-argument-similarity values, and any reference to parallel language versions. The judge returns integer scores from 1 to 10 for argumentative development, engagement, final argument quality, and overall quality. Twenty randomly selected sessions are rescored; Pearson test--retest reliability ranges from $0.84$ to $1.00$, and exact agreement ranges from 85\% to 100\% across dimensions.

\begin{PromptBlock}
Read the motion and the complete eight-turn debate. Score each dimension from 1 to 10.
- argumentative_development: whether later turns add, refine, or productively contest reasons
- engagement: whether speakers address the opposing side and the debate's unresolved points
- final_argument_quality: strength and completeness of the arguments available by the end
- overall_quality: holistic quality of the debate as a deliberative exchange
Return JSON only:
{"argumentative_development": 1-10,
 "engagement": 1-10,
 "final_argument_quality": 1-10,
 "overall_quality": 1-10,
 "brief_rationale": "..."}
\end{PromptBlock}
The displayed schema normalizes the English and Chinese wrapper text while preserving the scored dimensions and blinding constraints.

\subsection{Association Between Process and Outcomes}
We first ask whether the process score moves with judged quality at all. All eight baseline language--dimension Spearman correlations are negative. The strongest cell is English argumentative development, $\rho=-0.29$ with a 95\% motion-cluster bootstrap CI of $[-0.47,-0.08]$. The judge uses only three to six distinct integer values in some language--dimension cells, so tied ranks attenuate the observed associations. Table~\ref{tab:process-outcome-associations} reports the corresponding regression estimates.

\begin{table*}[t]
\centering
\small
\begin{tabular}{lcc}
\toprule
Analysis & Estimate & Uncertainty \\
\midrule
EN baseline Spearman: development & $-0.29$ & $[-0.47,-0.08]$ \\
Motion-FE: development per $+0.01$ similarity & $-0.16$ & $p<0.01$ \\
Motion-FE: final quality per $+0.01$ similarity & $-0.10$ & $p<0.01$ \\
\bottomrule
\end{tabular}
\caption{Selected process--outcome associations.}
\label{tab:process-outcome-associations}
\end{table*}

The motion-fixed-effects regressions use all 284 sessions and motion-clustered standard errors. The estimates show convergent validity but not determinism: sessions with similar process scores can receive different outcome scores, and the process metric should not replace direct quality evaluation.

\subsection{Intervention Payoff by Language}
A negative association does not by itself imply that lowering the score improves the debate. We therefore compare what the diversity prompt buys in each language, on the process score and on the judged outcomes jointly (Table~\ref{tab:quality-intervention-did}).

\begin{table*}[t]
\centering
\small
\begin{tabular}{lrrr}
\toprule
Measure & English & Chinese & Difference-in-differences (EN $-$ ZH) \\
\midrule
Process score, BGE-M3 & $-0.023^{*}$ & $-0.020^{*}$ & $-0.003$ $[-0.012,+0.005]$ \\
Argumentative development & $+0.90^{*}$ & $+0.96^{*}$ & $-0.06$ $[-0.27,+0.17]$ \\
Engagement & $+0.70^{*}$ & $+0.10$ & $+0.61$ $[+0.37,+0.86]$ \\
Final argument quality & $+0.76^{*}$ & $+0.27$ & $+0.49$ $[+0.13,+0.87]$ \\
\bottomrule
\end{tabular}
\caption{Diversity-minus-baseline changes in the process score and judged outcomes. Asterisks mark paired 95\% CIs excluding zero.}
\label{tab:quality-intervention-did}
\end{table*}

The process-score reduction is statistically indistinguishable across languages, and development improves in both. Engagement and final-quality gains are significant only in English, and the cross-language differences in those gains are themselves significant. English diversity sessions are near the development ceiling (mean $9.03$, SD $0.17$), while diversity transcripts lengthen more in Chinese ($1.30\times$) than English ($1.24\times$); neither pattern explains an artificial English advantage through additional score range or text length. Baseline Chinese engagement is $0.25$ points lower than English ($[-0.48,-0.07]$), but a single bilingual judge may have language-specific scale tendencies, so we do not treat these scores as a direct ranking of language quality.

\subsection{Preliminary Early-Stopping Probe}
For 20 motions, the same judge scores 4-, 6-, and 8-turn prefixes. Mean final-quality scores are already approximately $8.2/10$ at turn~4, but the 4-to-8-turn gains are indistinguishable from zero at this judge's ceiling-limited resolution, and early prior-argument similarity does not reliably predict the later gain. The confidence intervals are wide. These results are compatible with diminishing returns but do not validate an early-stopping rule; a stopping policy would require more motions, multiple judges, task-specific utility and error costs, and prospective evaluation.

\section{Responsible Research and Reproducibility Details}
\label{app:responsible-research}

\subsection{Risks, Scope, and Intended Use}
\label{app:risks-intended-use}

This work studies generated debate transcripts as research artifacts; it does not deploy a debate system, make claims about human debaters, or rank human languages or cultures. A potential risk is that prior-argument similarity could be over-interpreted as an item-level quality score, a direct repetition label, or evidence of a property of human Chinese or English argumentation. We mitigate this risk by defining the metric as an aggregate diagnostic, reporting manual calibration limits, and framing the Chinese--English result as a model--language process effect under the tested pipeline.

The debate motions include political, social, and ethical topics. Generated turns may therefore contain persuasive, unsupported, or culturally sensitive claims even though the prompts prohibit fabricated statistics and the analysis treats outputs only as model-generated text. The intended use of the released methods and derived statistics is research on multilingual LLM evaluation and debate-process diagnostics. They should not be used as a standalone moderation tool, debate judge, language ranking, or user-facing deliberation system without additional human evaluation and deployment-specific risk review.

\subsection{Artifacts, Licenses, and Data Content}
\label{app:artifact-license-details}

The study uses existing artifacts and creates derived research artifacts. Existing artifacts include the WUDC motion archive~\citep{tokyodebate_wudc_motions}, provider-hosted debate agents~\citep{openai2026gpt55,anthropic2026opus47,deepseek2025v32,mistral2025large3}, the Llama-4-Scout-17B-16E-Instruct extractor~\citep{meta2025llama4scout}, multilingual embedding models~\citep{chen2024bgem3,wang2024multilinguale5,feng2022labse}, the BGE reranker used for tail rescoring~\citep{baai2026bgererankerv2m3}, and vLLM for local extractor serving~\citep{kwon2023vllm}. The created artifacts are generated debate transcripts, translated motions, argument-unit extractions, similarity scores, diagnostic tables, and analysis scripts.

We use the above artifacts for research evaluation under their respective access terms and licenses. The provider-hosted models are accessed through APIs rather than redistributed. The locally used Llama extractor is used under the Llama 4 Community License; BGE-M3 and multilingual-E5-large are distributed under MIT-style model licenses on Hugging Face; LaBSE and BGE-reranker-v2-m3 are distributed under Apache-2.0 licenses on Hugging Face. Any public release of derived transcripts, code, or processed tables should retain artifact citations, avoid redistributing model weights, and include the accompanying documentation of languages, prompts, model identifiers, and intended research-only use.

The source motions are public debate motions and do not contain private participant data. The generated transcripts are synthetic model outputs. We do not intentionally collect personally identifying information, and no user conversations or private documents are used. Because some motions concern sensitive public issues, generated outputs may mention public groups, countries, or institutions and may contain controversial claims; these outputs are retained only as research data for aggregate analysis.

\end{document}